\documentclass[lettersize,journal]{IEEEtran}
\usepackage{amsmath,amsfonts}
\usepackage{algorithmic}
\usepackage{algorithm}
\usepackage{array}
\usepackage[caption=false,font=normalsize,labelfont=sf,textfont=sf]{subfig}
\usepackage{textcomp}
\usepackage{stfloats}
\usepackage{url}
\usepackage{verbatim}
\usepackage{graphicx}
\usepackage{cite}
\usepackage{hyperref}
\usepackage{multirow}
\usepackage{booktabs}
\usepackage{makecell}
\usepackage{xcolor}
\usepackage{fontawesome5}

\newcommand{\bT}{\mathbf{T}}

\makeatletter
\DeclareRobustCommand\onedot{\futurelet\@let@token\@onedot}
\def\@onedot{\ifx\@let@token.\else.\null\fi\xspace}

\makeatother

\newcommand{\figref}[1]{Fig.~\ref{#1}}
\newcommand{\secref}[1]{Section~\ref{#1}}

\renewcommand{\eqref}[1]{Eq.~\ref{#1}}
\newcommand{\tabref}[1]{Table~\ref{#1}}

\newcommand{\boldparagraph}[1]{\vspace{0.2cm}\noindent{\bf #1:} }

\newif\ifcomment
\commenttrue
\ifcomment
	\newcommand{\ag}[1]{ \noindent {\color{red} {\bf Andreas:} {#1}} }
	\newcommand{\yl}[1]{ \noindent {\color{cyan} {\bf YL:} {#1}} }
        
        \newcommand{\czw}[1]{ \noindent {\color{brown} {\bf Chengzhen:} {#1}} }
\else
	\newcommand{\ag}[1]{}
	\newcommand{\yl}[1]{}
        \newcommand{\czw}[1]{}
\fi

\newcommand{\gsplat}{\textit{gsplat}}

\begin{document}

\title{GSCodec Studio: A Modular Framework for Gaussian Splat Compression}

\author{Sicheng Li*, Chengzhen Wu*, Hao Li, Xiang Gao, Yiyi Liao\textsuperscript{\textdagger}, Lu Yu\textsuperscript{\textdagger},~\IEEEmembership{Senior Member,~IEEE}
\thanks{Sicheng Li, Chengzhen Wu, Hao Li, Xiang Gao, Yiyi Liao, Lu Yu are with Zhejiang University. Email: \{jasonlisicheng, chengzhenwu, 22231096, xiangg, yiyi.liao, yul\}@zju.edu.cn}%
\thanks{*Sicheng Li and Chengzhen Wu contributed equally to this work.}
\thanks{\textsuperscript{\textdagger}Corresponding authors: Yiyi Liao and Lu Yu.}
}

\markboth{Journal of \LaTeX\ Class Files,~Vol.~14, No.~8, August~2021}%
{Shell \MakeLowercase{\textit{et al.}}: A Sample Article Using IEEEtran.cls for IEEE Journals}

\maketitle

\begin{abstract}
3D Gaussian Splatting and its extension to 4D dynamic scenes enable photorealistic, real-time rendering from real-world captures, positioning Gaussian Splats (GS) as a promising format for next-generation immersive media. However, their high storage requirements pose significant challenges for practical use in sharing, transmission, and storage. Despite various studies exploring GS compression from different perspectives, these efforts remain scattered across separate repositories, complicating benchmarking and the integration of best practices. To address this gap, we present GSCodec Studio, a unified and modular framework for GS reconstruction, compression, and rendering. The framework incorporates a diverse set of 3D/4D GS reconstruction methods and GS compression techniques as modular components, facilitating flexible combinations and comprehensive comparisons. By integrating best practices from community research and our own explorations, GSCodec Studio supports the development of compact representation and compression solutions for static and dynamic Gaussian Splats—namely our Static and Dynamic GSCodec—achieving competitive rate-distortion performance in static and dynamic GS compression. The code for our framework is publicly available at \href{https://github.com/JasonLSC/GSCodec_Studio}{https://github.com/JasonLSC/GSCodec\_Studio}, to advance the research on Gaussian Splats compression.
\end{abstract}

\begin{IEEEkeywords}
Novel View Synthesis, Volumetric Video Compression, Gaussian Splatting Compression
\end{IEEEkeywords}

\section{Introduction}
\label{sec:intro}
\IEEEPARstart{V}{olumetric} video, as the next generation of visual media format, allows users to freely view dynamic 3D content from arbitrary viewpoints, showing significant potential in entertainment, education, and other fields.
With the photorealistic rendering and real-time rendering speed, 3D Gaussian Splatting (3DGS)~\cite{kerbl3Dgaussians} has quickly emerged as a significant representation for volumetric video. 
A series of follow-up works have extended Gaussian Splatting to dynamic modeling~\cite{luiten2023dynamic3DG,Wu20244dgaussiansplatting,yang2023deformable3dgs,yang2023gs4d,li2024spacetimegaussian,kratimenos2024dynmf}, enabling highly realistic content viewing at arbitrary timestamps and viewpoints from real-world captures.
Together, the emergence of 3D and 4D Gaussian Splatting, offering users compelling immersion and interactivity, has reshaped the landscape of volumetric video, and even the future of virtual reality and augmented reality.

Despite the rapid progress, a key drawback of these methods is the large storage requirement of the representation, Gaussian Splats (GS)\footnote{We refer to the data used in 3DGS/4DGS as \textit{Gaussian Splats.}}, which limits efficient transmission and storage as an interactive media format. 
Efficient GS compression is therefore essential to reduce storage needs while maintaining high visual quality, especially for dynamic scenes where the temporal dimension further increases data volume.

Notable progress has been made in 3D/4D GS reconstruction and compression. However, most efforts are either developed independently or not open-sourced, lacking a unified platform for comprehensive benchmarking and comparison. Specifically, no platform currently examines how to effectively integrate representation design and compression, particularly for 4D representations. This gap limits the discovery of best practices for GS compression and hinders further algorithm development. Therefore, a unified platform is needed to connect 3D/4D GS reconstruction, compression, and rendering, enabling research progress and codec standard prototyping.

\begin{figure*}[t]
    \centering
    \includegraphics[width=0.95\linewidth]{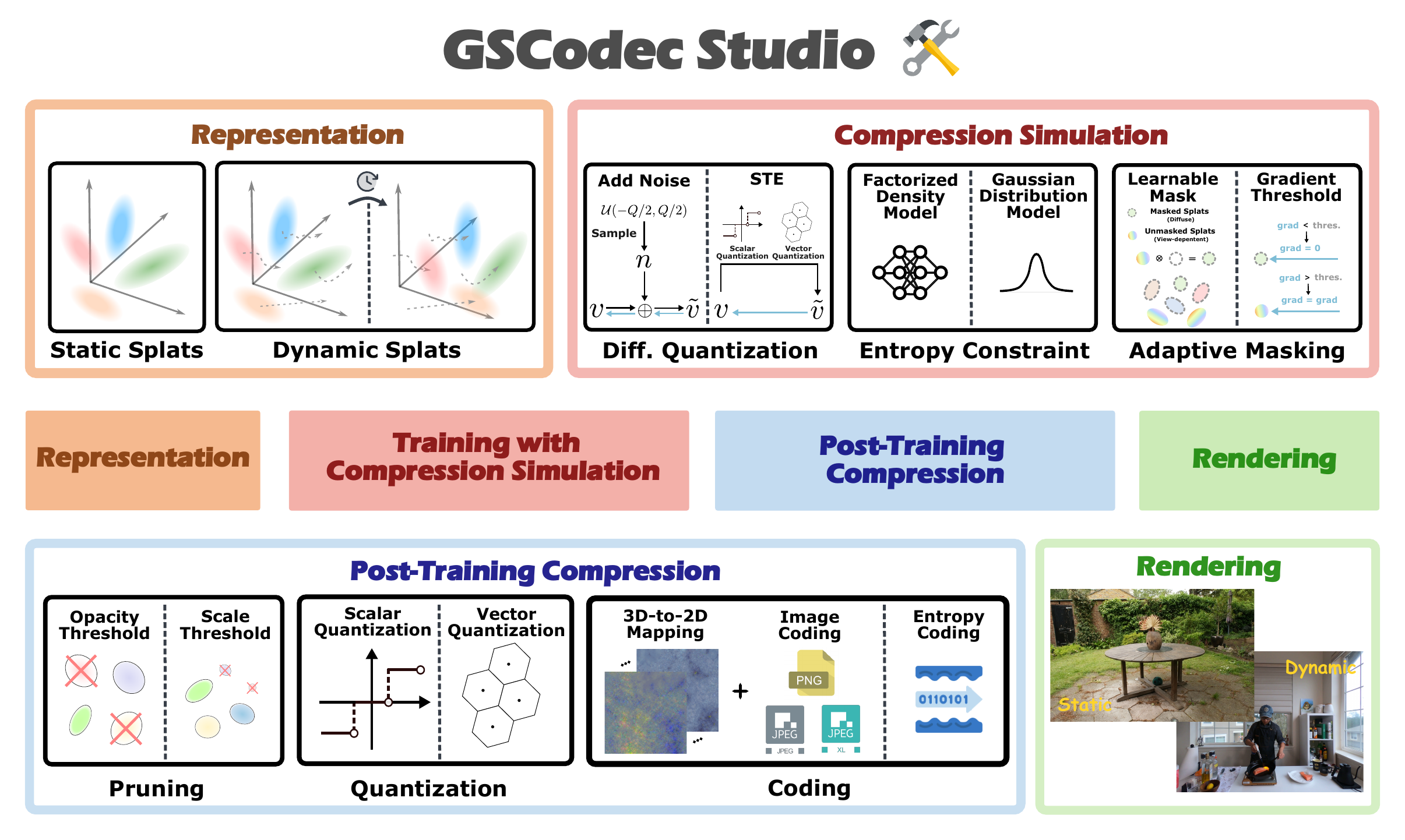}
    \caption{\textbf{GSCodec Studio} is a modular framework that unifies various Gaussian Splatting reconstruction and compression techniques to push the frontier of compression efficiency of Gaussian Splatting. }
    \label{fig:teaser}
\end{figure*}

To advance research in this field, we propose \textbf{GSCodec Studio}, an integrated and modular framework for 3D/4D GS reconstruction, compression, and rendering, as shown in \figref{fig:teaser}. It incorporates a variety of 3D/4D GS representations and reconstruction methods, along with a range of effective compression techniques inspired by both the image/video compression community and recent advances in Gaussian Splat.
For \textit{representation}, we offer several static and dynamic GS representations, including designs that consider compactness.
In terms of \textit{compression}, we first categorize existing GS compression techniques into two main branches: training with compression simulations and post-training compression, depending on whether they need to integrate with GS training. We then summarize several effective and orthogonal representative technical paths under these two branches. Subsequently, we reimplement representative methods for each of these paths within our framework and also introduce some improved new methods.
For \textit{rendering}, our framework supports real-time rendering of both static and dynamic scenes, and provides a real-time 4D player that enables interactive free-viewpoint viewing and free playback.

Our framework integrates a range of existing GS compression techniques, enabling comparison of different methods in a controlled environment. Through extensive ablation studies, we assess the impact of various design choices, providing insights for identifying best practices. By incorporating best practices, we introduce promising static and dynamic GS compression solutions, Static and Dynamic GSCodec, which achieve competitive performance in 3D/4D GS compression.

We summarize our contribution as follows. i) We provide a modular, open-source platform for GS compression research, integrating various static and dynamic representation forms along with representative compression techniques.
ii) 
To demonstrate the utility of our platform for efficient compression and method development, we introduce two representative compositional methods, Static GSCodec and Dynamic GSCodec, achieving advanced rate-distortion performance in 3D/4D Gaussian Splat compression. 
iii) Our extensive experiments offer valuable insights and highlight key challenges of GS coding. A notable finding is that entropy constraints lead to a more compact parameter distribution, significantly improving the compressibility of Gaussian Splats. Meanwhile, efficiently compressing the spatial positions of Gaussian Splats remains a major challenge, warranting further investigation.

\section{Related Work}
\label{sec:related}
\boldparagraph{Gaussian Splatting for 3D/4D Reconstruction}
Gaussian splatting~\cite{kerbl3Dgaussians,bao20253dgssurvey} has recently received great attention with many works proposed to improve its performance~\cite{charatan23pixelsplat, Yu2024MipSplatting, kheradmand20243dgsmcmc, chen2024mvsplat, Huang20242DGS} and expand its application scope in other fields of vision, such as digital humans~\cite{hu2024gaussianavatar,qian2024gaussianavatars,saito2024relightable,zheng2024gpsgaussian,zhang2025humanrefgs}, generative modeling~\cite{tang2024dreamgaussian, xu2024grm}, SLAM~\cite{yan2024GSSLAM, keetha2024splatam}, telepresence~\cite{tu2024telealoha}, autonomous driving~\cite{cui2025streetsurfgs}, physics simulation~\cite{xie2024physgaussian, jiang2024vrgs}, and robotics~\cite{abou2024physically}.
A significant portion of research focuses on extending 3DGS to model dynamic scenes~\cite{luiten2023dynamic3dgs, Wu20244dgaussiansplatting, sun20243dgstream, guo2024motion3dgs, li2025frpgs}. Dynamic 3D Gaussians~\cite{luiten2023dynamic3dgs} is the first to extend 3DGS to 4D by storing the time-dependent parameters individually at each timestamp, leading to large memory consumption. The following methods propose various dynamic representations by sharing parameters across space and/or time by using feature planes~\cite{Wu20244dgaussiansplatting}, MLPs~\cite{yang2023deformable3dgs}, hash grids~\cite{sun20243dgstream}, sparse control points~\cite{huang2024scgs}, motion basis~\cite{kratimenos2024dynmf, wang2024shapeofmotion}, or 4D Gaussians~\cite{yang2023gs4d,duan20244drotorgs}.
Our framework supports static Gaussian Splats with different shading mechanisms and integrates representative dynamic Gaussian Splats based on their motion representation and opacity formulation.

\boldparagraph{Gaussian Splats Compression}
Due to the large storage cost, a series of works attempt to address GS compression. 
Several methods investigate compact design of the representation~\cite{lu2024scaffold, ververas2024sags}.
Pruning~\cite{papantonakis2024reducing,fan2023lightgaussian,lee2024Compact3DGaussian,lee2024C3DGS, fang2024minisplatting} is an effective strategy for removing GS points that contribute little to rendering quality, significantly reducing storage and improving rendering speed. Quantization~\cite{papantonakis2024reducing,navaneet2023compact3d,niedermayr2024compressed3dgs,chen2024hac,morgenstern2023sogs,xie2025mesongs} is a highly efficient method for decreasing the bit usage of each Gaussian point's attributes, including both vector and scalar quantization. 
Entropy coding based on a learned distribution~\cite{chen2024hac,wang2024contextgs} is also effective for GS compression.
In contrast to methods incorporating rate-distortion optimization during GS training~\cite{wang2024rdo3dgs,chen2024hac,liu2024compgs}, Self-organization GS~\cite{morgenstern2023sogs} proposes to sort 3D Gaussians onto a 2D image and uses an off-the-shelf 2D codec.
Compared with 3D GS compression, there are fewer investigations in 4D GS compression~\cite{Jiang_2024_CVPR,jiang2024robust,wang2024v3, lee2024C3DGS, bae2024EmbedD3dgs, zhang2024mega, kwak2025modec}. Most dynamic methods share similar principles with static GS compression, while some also consider temporal correspondence to remove redundancy in the temporal domain~\cite{girish2024queen, zhang2025evolvinggs,wang2024v3}.

\boldparagraph{Frameworks and Tools for Gaussian Splatting}
Several Gaussian Splatting frameworks and tools are currently available online. \gsplat~\cite{ye2024gsplatopensourcelibrarygaussian} is an open-source, efficient, and user-friendly library with several algorithmic enhancements and features. GauStudio~\cite{ye2024gaustudio} is a modular framework consolidating various research efforts into a single code repository. EasyVolCap~\cite{xu2023easyvolcap} is a library for accelerating neural volumetric video research with its own dynamic Gaussian Splatting implementation. Pointrix is a differentiable point-based rendering library focusing on Gaussian Splatting. The well-known NeRF development framework NeRFStudio~\cite{nerfstudio} has also released its own GS implementation called Splatfacto, a blend of different Gaussian splatting methodologies. These pioneer works deeply inspire our framework. Distinct from them, ours focuses more on advancements related to GS compression, including compact static and dynamic representations, diverse compression strategies, and the integration of efficient data compression methods from the broader compression community.

\section{Framework Design}
\begin{figure*}[t]
    \centering
    \includegraphics[width=0.95\linewidth]{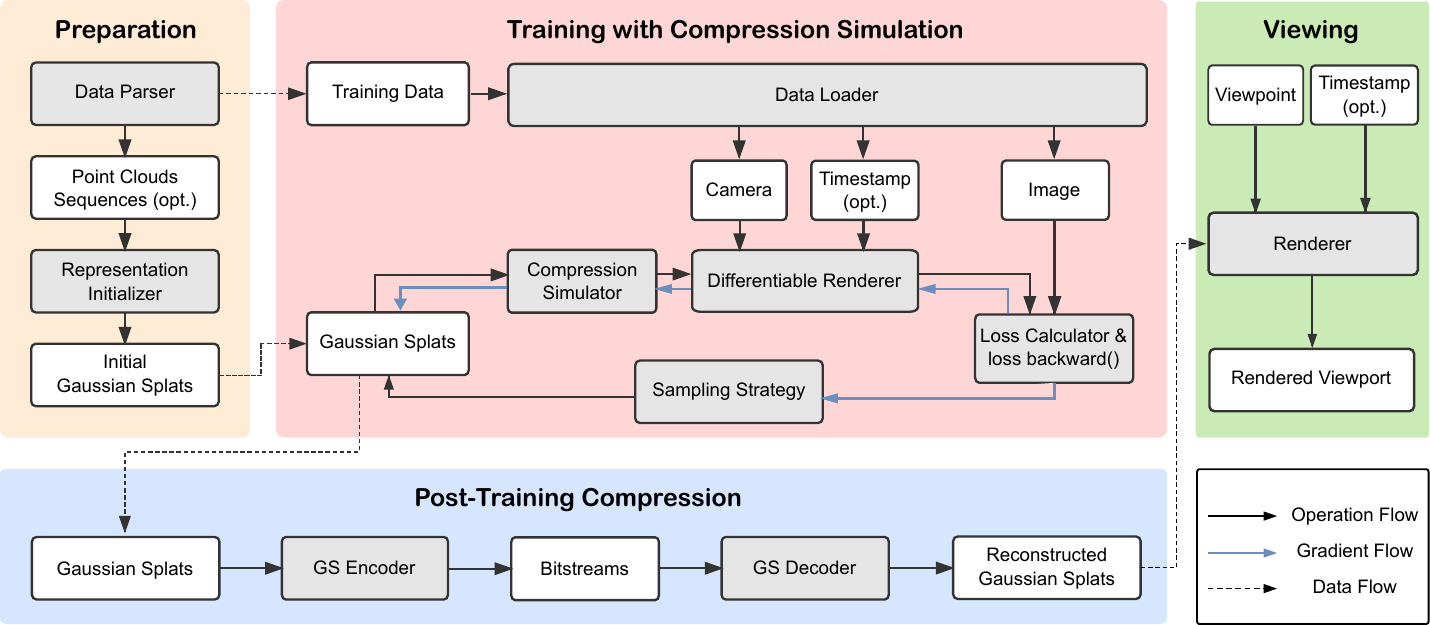}
    \caption{
        \textbf{Abstraction and Code Design of the Framework.} The framework consists of four main parts: Preparation (beige), Training with Compression Simulation (pink), Post-Training Compression (blue), and Viewing (green). The preparation stage processes input data and initializes scene representations. The training with compression simulation stage handles the training of Gaussian Splats and introduces compression simulation during the training process, enabling the generation of more compact and compressible Gaussian Splats. The post-training compression stage performs the actual encoding and decoding processes of Gaussian Splats. The viewing stage takes Gaussian Splats and the control signal (viewpoint and timestamp) as input and renders the corresponding viewport. Arrows indicate operation flow, data flow, and gradient flow between different components. The operation class is shown in gray, and the data class is in white.
    }
    \label{fig:framework}
\end{figure*}
The goal of GSCodec Studio is to provide a comprehensive framework that integrates various effective Gaussian Splats compression techniques, allowing users to experiment with different combinations and compare their strengths and weaknesses. To help readers better understand our framework, we present its design philosophy, core modules, and the relationships between them, along with the illustration in \figref{fig:framework}.

The framework is composed of four key stages: 1) Preparation, 2) Training with Compression Simulation, 3) Post-Training Compression, and 4) Viewing, as illustrated in \figref{fig:framework}. 

The \textit{Preparation} stage handles parsing Colmap data and initializing static or dynamic splats representations. It provides the necessary supervision data for the training process and sets up the scene representations to be used during training.

 During the \textit{Training with Compression Simulation} stage, training data is processed via a dataloader that prepares the data for each iteration. In each iteration, the splats are subjected to compression simulation before being rendered, followed by loss calculation and gradient backpropagation. This integration of compression simulation helps in training the splats to be more compact and compressible.

Once the training is complete, the splats undergo \textit{Post-Training Compression} using a GS encoder and decoder. The encoder applies techniques such as pruning, quantization, sorting, and image coding to generate a compact bitstream. The decoder then reconstructs the splats through image decoding and dequantization, producing the final decoded splats.

The \textit{Viewing} stage renders the current scene by synthesizing the corresponding view window based on the viewpoint and timestamp information (for dynamic scenes), which are determined through user interaction. This enables the real-time rendering of Gaussian Splats based on the user’s perspective.

In summary, GSCodec Studio provides a modular framework organized into four key stages: Preparation, Training with Compression Simulation, Post-Training Compression, and Viewing. Within these stages, a set of core components work together to realize the modularity of the framework, offering flexibility for exploring rate-distortion trade-offs in GS compression. The following section details these core components and their specific functionalities.

\section{Core Components}
Our GSCodec Studio supports high-quality representation and efficient compression of both static and dynamic scenes.
To achieve this, GSCodec Studio includes the following core components: static representation~(\secref{sec:static}), dynamic representation~(\secref{sec:dynamic}), training with compression simulation~(\secref{sec:training_simulation}), post-training compression~(\secref{sec:compression}), and a real-time 4D player for viewing~(\secref{sec:player}).

\subsection{Static Representation}
\label{sec:static}
Our framework supports two types of static Gaussian Splatting: one is the vanilla 3DGS, and the other is a more compact variant we call feature-based 3DGS. The vanilla 3DGS uses the existing efficient implementation in \gsplat. In feature-based 3DGS, spherical harmonics (SH) coefficients are replaced with lower-dimensional learned features. The rendered features in 2D space are then mapped to color using a small MLP for shading. This approach is similar to  MobileNeRF~\cite{chen2022mobilenerf}, where a rendered feature image is converted into an RGB image via a shallow MLP. 

\begin{figure}[t]
    \centering
    \includegraphics[width=0.9\linewidth]{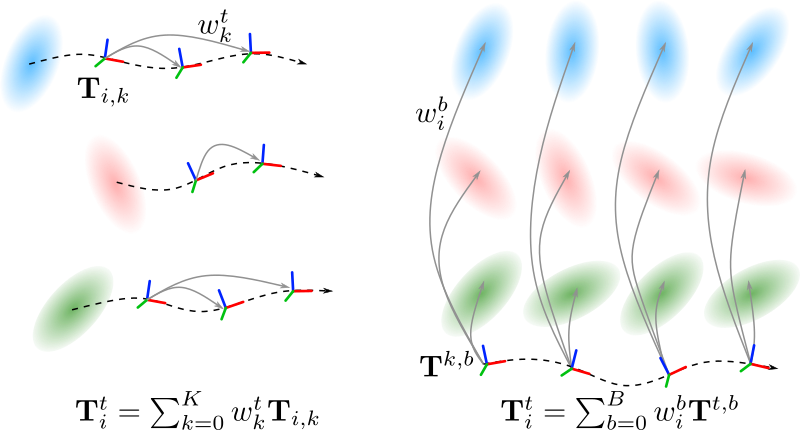}
    \caption{\textbf{Motion Representations} of dynamic Gaussians. We implement two representative representations for the time-variant motion $\bT_i^t \in SE(3)$ with $i$ denoting the Gaussian index and $t$ the timestamps. \textit{Left}: Motion parameters are shared over the time dimension for each individual 3D Gaussian. \textit{Right}: Motion parameters are shared across all 3D Gaussians.}
    \label{fig:dynamic_rep}
\end{figure}

\subsection{Dynamic Representation}
\label{sec:dynamic}
To model dynamic scenes, existing 4DGS methods use time-varying 3D Gaussians, with some attributes shared across time (e.g., SH coefficients) and some varying (e.g., transformation and/or opacity). 
In GSCodec Studio, we focus on two main design aspects for such 4D representations: motion representation and opacity formulation, and we provide modular implementations.

\boldparagraph{Motion Representation} To model content movement over time,  positions and rotations of GS are typically time-varying attributes for dynamic scenes. 
Modeling these attributes independently, as in \cite{luiten2023dynamic3dgs}, is memory-intensive. To address this, recent 4DGS methods reduce memory use by sharing certain parameters across time or space~\cite{yang2023gs4d, li2024spacetimegaussian, wang2024shapeofmotion, kratimenos2024dynmf}, as illustrated in \figref{fig:dynamic_rep}. For instance, motion can be parameterized using methods such as an MLP, triplane, or motion basis. We implement two representative motion representations: 1) a polynomial-based trajectory representation~\cite{li2024spacetimegaussian}, which reduces parameters by sharing motion parameters over the \textit{time dimension for each individual 3D Gaussian}, see \figref{fig:dynamic_rep} left; and 2) a method using learnable bases and coefficients to characterize motion trajectories~\cite{wang2024shapeofmotion, kratimenos2024dynmf}, where the basis is shared across \textit{all 3D Gaussians} as shown in \figref{fig:dynamic_rep} right. Note that MLP and triplane-based motion representations are other alternatives for parameter sharing over all Gaussians and could easily be integrated into our framework.

\boldparagraph{Opacity Formulation}
We investigate both, time-independent and time-varying, opacity. The time-varying opacity allows it to model the appearance and disappearance of certain substances, such as the ignition and extinction of a flame. Like existing works~\cite{yang2023gs4d, duan20244drotorgs, li2024spacetimegaussian}, we use a temporal radial kernel function (Gaussian function) to parameterize the change of opacity.  By setting an opacity threshold, we can determine the lifespan of the Gaussian points.  

\subsection{Training with Compression Simulation}
\label{sec:training_simulation}
Learning a 3D representation from multi-view 2D captures is an ill-posed problem, yielding multiple solutions with similar rendering quality but varying significantly in compressed size. Without regularization during training, the solution may be suboptimal for compression.
To address this, we integrate compression simulation into the training process, enabling a rate-distortion optimal representation. This approach yields better rate-distortion performance in actual compression. Our training-time compression simulation includes three components: differentiable quantization, entropy constraint, and adaptive masking.

\boldparagraph{Differentiable Quantization}
Quantization is an effective and commonly used methodology in various GS compression solutions~\cite{niedermayr2024compressed3dgs, fan2023lightgaussian, navaneet2023compact3d, wang2024rdo3dgs}. In the existing practice in the compression of \gsplat, both scalar and vector quantization are involved, but only in a post-processing form.  Introducing quantization effects during training can reduce the accuracy loss from post-processing compression or enable lower bit-width compression with an acceptable quality loss.
This paradigm is widely applied in neural network compression~\cite{nagel2021nnquantization} and learning-based image compression~\cite{balle2018hyperprior, begaint2020compressai, ascenso2023jpegai}.
Since naive quantization is non-differentiable, approximate differentiable quantization methods have been proposed to enable its use in training. Our framework implements two such approaches: uniform noise approximation and the Straight-Through Estimator (STE).
\\
\noindent 1) \textit{Uniform Noise Approximation}: This method is suitable for approximating scalar quantization. Given the step size~$Q_s$ of scalar quantization, we add noise~$n$, which follows a uniform distribution~$U[-Q_s/2, Q_s/2]$, to the elements to be quantized~$v$, simulating the possible perturbations in actual quantization. 
\begin{equation}
    \tilde{v} = v + n, \quad n \in U[-Q_s/2, Q_s/2]
\end{equation}
where $\tilde{v}$ means the values added with noise.
\\
\noindent 2) \textit{Straight-Through Estimator}:
STE-based quantization supports both scalar and vector quantization. During forward inference, it performs the original operation, while in the backward pass, it assumes a gradient of 1, passing the received gradient unchanged to preceding nodes in the computation graph.
\begin{align}
    \text{Forward:} & \quad \tilde{v} = Q(v) \\
    \text{Backward:} & \quad \frac{\partial \mathcal{L}}{\partial v} = \frac{\partial \mathcal{L}}{\partial \tilde{v}} \cdot \frac{\partial \tilde{v}}{\partial v} = \frac{\partial \mathcal{L}}{\partial \tilde{v}} \cdot 1
\end{align}
\noindent{\textbf{Entropy Constraint:}}
Inspired by learning-based image compression, we can introduce an entropy constraint of attributes to be compressed into the training objective to achieve joint optimization of rate and distortion. This allows us to find a more optimal representation from the rate-distortion perspective and helps in test-time compression.
\begin{figure}[t]
    \centering
    \includegraphics[width=\linewidth]{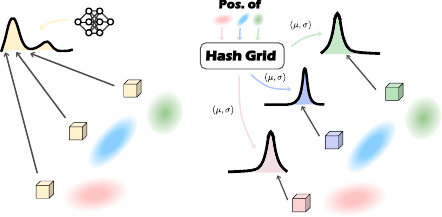}
    \caption{\textbf{Entropy models}. We illustrate the probability estimation for one channel across multiple Gaussian Splats. \textit{Left:} Factorized density model uses a shallow MLP to estimate a non-parametric distribution shared for all Gaussian Splats. \textit{Right:} Per-point Gaussian distribution model proposed in HAC learns a per-point Gaussian distribution for each Gaussian Splat.}
    \label{fig:entropy_model}
\end{figure}
The next question is how to derive the entropy of attribute values $v_i$. The key is to incorporate a probability estimator for the attributes. In learning-based compression, the probability estimator $F$ takes attribute values as input and outputs its cumulative probability density (CDF) under the assumed distribution. Given a quantized value $\tilde{v}_i$ and its corresponding quantization step $Q_s$, we can then obtain the probability $P(\tilde{v}_i)$ corresponding to the quantized value (also called a symbol) obtained through the quantization.
\begin{gather}
    P(\tilde{v}_i) = F(\tilde{v}_i + Q_s/2) - F(\tilde{v}_i - Q_s/2) \\
    \mathcal{L}_{entropy} = -\frac{1}{n}\sum_{i=1}^{n} \log[P(\tilde{v}_i)]
\end{gather}

Our framework implements two types of probability estimators from two different design principles. \\
\noindent \textit{1) Factorized density model}:
The factorized density model, proposed in~\cite{balle2018hyperprior}, is commonly used in learning-based image compression to estimate the probability distribution of latent variables $\boldsymbol{z}$ as side information. It learns a non-parametric distribution independently for each channel in the latent variables, assuming elements within a channel follow the same distribution. This distribution is parameterized by a shallow MLP, learned alongside the representation during training. When applied to Gaussian Splat attributes like scale, quaternion, and 0-order SH coefficients, we assume channel independence and use a shared learnable distribution within each channel, see \figref{fig:entropy_model} (left).\\
\noindent \textit{2) Per-point Gaussian distribution model}:
The second model is the Gaussian distribution approach with a learned hash grid context, proposed in HAC~\cite{chen2024hac}. HAC assumes each attribute element for encoding follows its \textit{own} Gaussian distribution, with the mean and standard deviation predicted by a hash grid and shallow MLP. This context is jointly optimized with the scene representation using a rate-distortion loss during training. Our implementation follows HAC, where we apply this Gaussian distribution model predicted from a hash grid to estimate the probability of attributes in the Gaussian Splats, see \figref{fig:entropy_model} (right).

\boldparagraph{Adaptive Masking}
Gaussian Splats may contain redundant attributes; for example, most points in a scene have diffuse color, not requiring anisotropic representation. To reduce the number of parameters, we train an adaptive mask to identify points that don’t need high-order SH coefficients. Similarly, in dynamic scenes, many points remain static, so an adaptive mask can also identify these during training, further enhancing representation compactness.
We implement two strategies for this purpose.\\
\noindent \textit{1) Point-wise learnable mask}:
The first strategy introduces a learnable, point-wise mask for each attribute, multiplied with the attribute during rendering. Initially, the mask has continuous values between 0 and 1, gradually becoming binary through training. At test time, a threshold converts it to a binary mask. A sparsity loss limits active points, while a KL loss ensures the active points' proportion aligns with the target value.\\
\noindent \textit{2) Gradient-based adaptive masking}:
The second strategy uses gradient-based adaptive masking. We initialize the attributes to be masked with zeros and, during backpropagation, set gradients below a predefined threshold to zero, retaining only high-magnitude gradients. This approach derives the adaptive mask based on gradients, effectively identifying points that require these attribute values.

\subsection{Post-Training Compression}
\label{sec:compression}
Our test-time compression extends beyond the implementation of \gsplat~\cite{morgenstern2023sogs, making_GS_more_smaller,torchpq} by integrating several best practices from current advanced explorations~\cite{morgenstern2023sogs,chen2024hac,lee2024Compact3DGaussian,navaneet2023compact3d}. The test-time compression pipeline includes pruning, 3D-to-2D mapping, quantization, image coding, and learned entropy coding.

\boldparagraph{Pruning}
At test time, even without training data, pruning can be performed using heuristics. Our framework includes criteria like opacity and scale thresholds for pruning. In Gaussian Splats trained with \gsplat~ and MCMC sampling~\cite{kheradmand20243dgsmcmc}, at least 5\% of points have opacity below this threshold, with negligible impact on quality metrics. Our framework also constructs a KDTree to identify and remove spatial outlier points.

\boldparagraph{Variable Bit-width Quantization}
The bit-width of quantization impacts compression: lower bit-widths improve compression ratios but may reduce rendering quality. Our framework mitigates this quality drop by integrating quantization-aware training. We support variable bit-width quantization, with 8-bit as the default and options for others such as 5- and 6-bit quantization. On the Tanks\&Temples dataset, PSNR values without quantization-aware training are 23.78dB and 22.78dB for 6- and 5-bit quantization. Only with 8-bit quantization-aware training, PSNR improves to 24.02dB and 23.65dB, yielding gains of 0.24dB and 0.87dB. Note that the experiments here only verify the impact of quantization, without involving other modules.

\newcommand{\enabled}[0]{{\color{yellow}\faLightbulb}}
\newcommand{\disabled}[0]{{\color{gray}\faLightbulb}}

\begin{table*}[t]
\centering
\small
\setlength{\tabcolsep}{3pt}
\caption{\textbf{GSCodec Studio modules \& Summary of GS compression methods}. We show all of the implemented compression modules in GSCodec Studio, covering most of the existing techniques for GS compression. This allows for straightforward implementations of existing static and dynamic GS compression methods in our framework. The last row illustrates Static GSCodec and Dynamic GSCodec, combining the best practices we discover after controlled experiments using our framework.}
\begin{tabular}{l|>{\centering\arraybackslash}c>{\centering\arraybackslash}c|>{\centering\arraybackslash}c>{\centering\arraybackslash}c|>{\centering\arraybackslash}c>{\centering\arraybackslash}c|c>{\centering\arraybackslash}c|c|c>{\centering\arraybackslash}c|cc}
\toprule
\multirow{2}{*}{\textbf{Methods}} & \multicolumn{2}{c|}{\textbf{Diff. Quant.}} & \multicolumn{2}{c|}{\textbf{Entropy Constr.}} & \multicolumn{2}{c|}{\textbf{Ada. Mask.}} &  \multicolumn{2}{c|}{\textbf{P.T. Quant.}} & \multirow{2}{*}{\textbf{Pruning}} & \multicolumn{2}{c|}{\textbf{Mapping}} & \multicolumn{2}{c}{\textbf{Coding}} \\

\cmidrule(l{2pt}r{2pt}){2-3} \cmidrule(l{2pt}r{2pt}){4-5} \cmidrule(l{2pt}r{2pt}){6-7} \cmidrule(l{2pt}r{2pt}){8-9} \cmidrule(l{2pt}r{2pt}){11-12} \cmidrule(l{2pt}r{2pt}){13-14}

 & \textbf{Noise} & \textbf{STE} & \textbf{Factor.} & \textbf{Gauss.} & \textbf{Learnable} & \textbf{Grad.} & \textbf{SQ} & \textbf{VQ} & & \textbf{Sort} & \textbf{Triplane} & \textbf{Img. C.} & \textbf{Learned C.}\\
\midrule

HAC~\cite{chen2024hac}             & \enabled & \disabled & \disabled & \enabled & \enabled & \disabled & \enabled & \disabled & \disabled & \disabled & \disabled & \disabled & \enabled \\
SOGS~\cite{morgenstern2023sogs}            & \disabled & \disabled & \disabled & \disabled & \disabled & \disabled & \enabled & \disabled & \disabled & \enabled & \disabled & \enabled & \disabled  \\
IGS~\cite{wu2024IGS}       & \enabled & \disabled & \disabled & \disabled & \disabled & \disabled & \enabled & \disabled & \disabled & \disabled & \enabled & \enabled & \disabled \\
CompGS~\cite{navaneet2023compact3d}       & \disabled & \enabled & \disabled & \disabled & \disabled & \disabled & \disabled & \enabled & \enabled & \disabled & \disabled & \disabled & \disabled \\
RDO-GS~\cite{wang2024rdo3dgs}   & \disabled & \enabled & \enabled & \disabled & \enabled & \disabled & \enabled & \enabled & \enabled & \disabled & \disabled & \disabled & \enabled \\
LightGaussian & \disabled & \enabled & \disabled & \disabled & \disabled & \disabled & \disabled & \enabled & \enabled & \disabled & \disabled & \disabled & \disabled \\

\midrule

CompactSTG & \disabled & \enabled & \disabled & \disabled & \enabled & \disabled & \enabled & \enabled & \enabled & \disabled & \disabled & \disabled & \disabled \\ 

\midrule
Ours (St.\&Dy.)   & \enabled & \disabled & \enabled & \disabled & \enabled & \disabled & \enabled & \enabled & \enabled & \enabled & \disabled & \enabled & \enabled \\

\bottomrule
\end{tabular}

\label{tab:comp_methodology}
\end{table*}

\boldparagraph{3D-to-2D Mapping \& Image Coding}
Although GS is an unstructured scene representation, strong correlations exist between the attributes of each point. The 3D-to-2D algorithm can transform these unordered 3D attributes into a smooth 2D image, enabling redundancy reduction through intra-prediction and transform coding. SOGS first introduced this approach, using Parallel Linear Assignment Sorting (PLAS) to create a 2D grid with local homogeneity, followed by standard image coding for compression. In \gsplat, 3D-to-2D mapping is applied post-training, with PNG for lossless coding. We follow the implementation in \gsplat~and extend this to compress dynamic GS, where dynamic attributes, like static ones, are arranged as 2D images and further compressed using image coding.

\boldparagraph{Entropy Coding from Learned Distribution}
With entropy constraint enabled in compression simulation, we obtain a learned probability distribution for each constrained attribute, regardless of the entropy model used. Given the learned distribution, quantized values, and steps, we can determine the probability of each quantized value (i.e., symbol) and perform entropy coding accordingly.

We integrate the Constriction library~\cite{bamler2022constriction} for entropy coding, which provides a set of entropy coding algorithms optimized for ease of use, compression performance, and computational efficiency. Specifically, we employ Asymmetric Numeral Systems (ANS)~\cite{duda2009ans} from the Constriction library as our entropy coder, a widely used method in learning-based image compression~\cite{balle2018hyperprior, cheng2020learned, begaint2020compressai, ascenso2023jpegai}.

\subsection{Real-Time 4D Player}
\label{sec:player}
\begin{figure}[ht]
    \centering
    \includegraphics[width=0.95\linewidth]{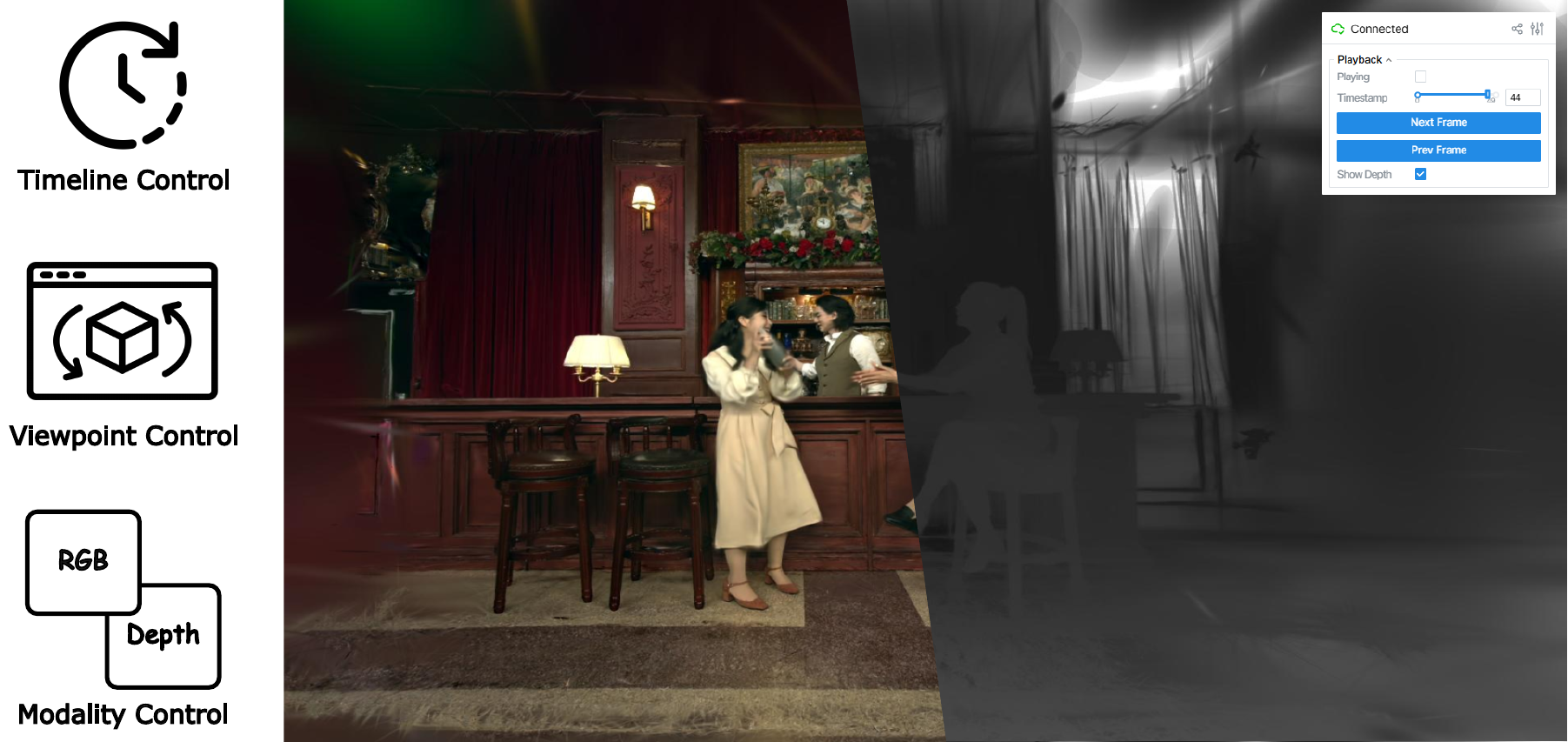}
    \caption{
    \textbf{Real-time 4D Player} supports free-viewpoint, free-time, and seamless switching between color/depth modalities for rendering.}
    \label{fig:player}
\end{figure}

In GSCodec Studio, we have developed a 4D player that supports real-time free-viewpoint rendering of dynamic scenes, as shown in \figref{fig:player}. This player allows users to view generated Dynamic Gaussian Splats and is useful for demo presentations and subjective quality assessment. Key features include: 1) real-time free-viewpoint rendering, 2) timeline controls for loop playback, frame stepping, and playback navigation, and 3) support for visualizing different modalities, such as RGB and depth images. This tool simplifies the examination of Gaussian Splats and enhances both demo presentations and quality evaluation.

\section{Representative Compositional Methods}
Our modular framework covers almost all the major GS compression methodologies currently available. This allows for easily implementing various approaches within our framework, as shown in  \tabref{tab:comp_methodology}. Our framework enables us to perform controlled comparisons of various compression techniques, allowing us to identify the best practices that can collectively achieve better compression rates. Based on these best practices, we introduce Static GSCodec and Dynamic GSCodec, using the same principle to compress both 3D and 4D GS representations. We briefly introduce the design choice of these two methods. 

\subsection{Static GSCodec}
The starting point of Static GSCodec is \gsplat~\cite{gsplat}, using raw 3DGS as the representation, MCMC as the sampling strategy, and employing quantization, 3D-to-2D mapping, and image coding for compression. Static GSCodec additionally incorporates differentiable quantization, entropy constraint, and adaptive masking during the training phase. Specifically, for operation choices, Static GSCodec selects noise approximation, uses a factorized density model as the entropy model, and applies a learnable mask for adaptive masking. Moreover, Static GSCodec supplements the post-training compression in \gsplat~with pruning and entropy coding from the learned distribution. If an attribute has an entropy constraint applied during training, we use entropy coding from the learned distribution for post-training compression. Otherwise, we use image coding.

\subsection{Dynamic GSCodec}
Dynamic GSCodec uses STG~\cite{li2024spacetimegaussian} as the dynamic scene representation, with polynomial-based motion representation, time-varying opacity formulation, and feature-based appearance shading. We follow STG's adaptive density control strategy, which first densifies and then prunes. It trains with compression simulation and shares the same module choices as Static GSCodec. We use pruning, scalar quantization, 3D-to-2D mapping, image coding, and entropy coding in post-training compression. In line with Static GSCodec, for attributes that did not undergo entropy constraint during training, we apply image coding, while for those subjected to entropy constraint, we use entropy coding based on the learned distribution. 

\section{Experiments}
We benchmark Static GSCodec and Dynamic GSCodec against leading static and dynamic Gaussian Splats compression methods. Additionally, we demonstrate the modularity of our framework through ablation studies.

\subsection{Evaluation Conditions}
We evaluate rate-distortion performance across all methods, reporting commonly used metrics for visual quality: PSNR, SSIM, and LPIPS. For rate measurement, we report the compressed file size in MegaBytes (MB) for static scenes and the average bitrate in Megabits per second (Mbps) for dynamic scenes.

\subsection{Static GS Compression Comparison}

\boldparagraph{Datasets and Baselines}
Experiments are conducted on the Tanks\&Temples, MipNeRF360, and DeepBlending datasets. Baseline methods include SOGS~\cite{morgenstern2023sogs}, HAC~\cite{chen2024hac}, IGS~\cite{wu2024IGS}, and \gsplat~compression~\cite{gsplat, ye2024gsplatopensourcelibrarygaussian}.

\boldparagraph{Settings}
For Static GSCodec, we use MCMC as the sampling strategy. The maximum number of splats is set to 1 million for the Tanks\&Temples and MipNeRF360 datasets, and 0.49 million for the Deepblending dataset. To explore different rate points along the rate-distortion curve, we adjust the weight $\lambda$ of the rate loss in the entropy constraint module, setting it to 0.01, 0.006, and 0.002.

\begin{figure*}[t]
    \centering
    \begin{tabular}{c@{\vspace{-0.25cm}}}
        \includegraphics[width=\textwidth]{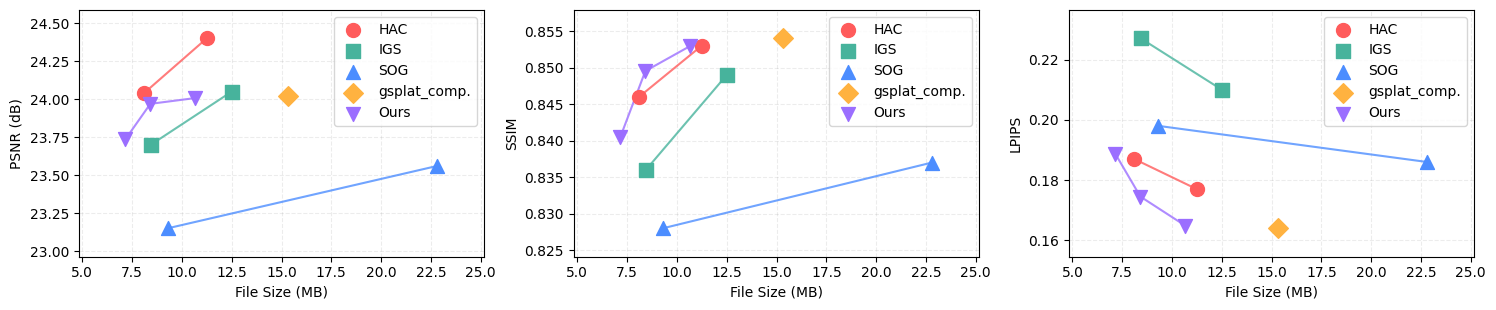} \\
        \includegraphics[width=\textwidth]{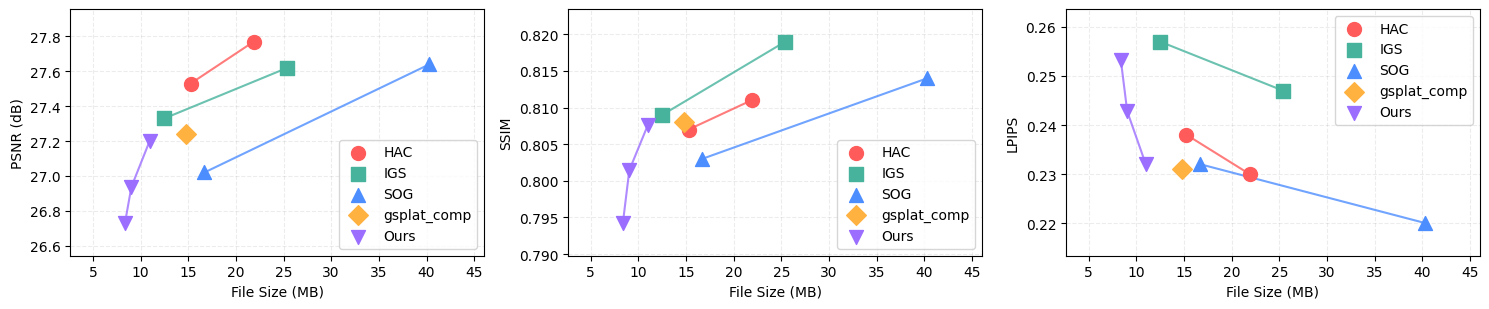} \\
        \includegraphics[width=\textwidth]{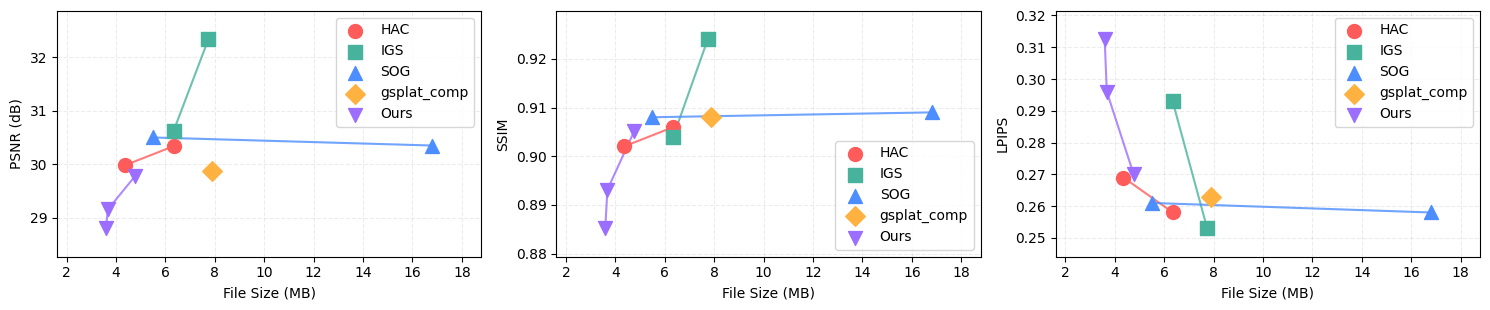}
    \end{tabular}
    \caption{\textbf{Rate-distortion comparisons on static datasets.} The three rows in this figure, from top to bottom, represent the results on Tanks\&Temples, MipNeRF360, and Deep Blending datasets.}
    \label{fig:static_rd_curve}
\end{figure*}
\begin{figure*}[t]
    \centering
    \begin{tabular}{c@{\vspace{-0.25cm}}}
        \includegraphics[width=\textwidth]{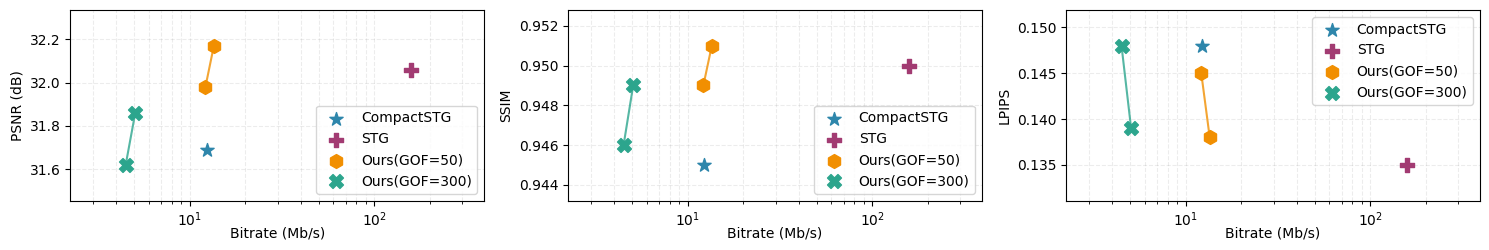} 
    \end{tabular}
    \caption{\textbf{Rate-distortion comparisons on the dynamic dataset}, Neural 3D Video dataset.}
    \label{fig:dyn_rd_curve}
\end{figure*}

\definecolor{tabfirst}{rgb}{1, 0.7, 0.7} 
\definecolor{tabsecond}{rgb}{1, 0.85, 0.7} 
\definecolor{tabthird}{rgb}{1, 1, 0.7} 

\begin{table*}[t]
\centering

\small 
\setlength{\tabcolsep}{4pt}
\caption{\textbf{The quantitative results of rate-distortion performance on static scenes.} The \colorbox{tabfirst}{best}, \colorbox{tabsecond}{second best}, and \colorbox{tabthird}{third best} results are highlighted.}
\begin{tabular}{l|cccc|cccc|cccc}
\toprule
\multirow{2}{*}{\textbf{Method}} & \multicolumn{4}{c|}{\textbf{Tanks and Temples}} & \multicolumn{4}{c|}{\textbf{MipNeRF 360}} & \multicolumn{4}{c}{\textbf{Deep Blending}} \\
\cmidrule{2-13}
& \textbf{PSNR}$\uparrow$ & \textbf{SSIM}$\uparrow$ & \textbf{LPIPS}$\downarrow$ & \textbf{Size}$\downarrow$ & \textbf{PSNR}$\uparrow$ & \textbf{SSIM}$\uparrow$ & \textbf{LPIPS}$\downarrow$ & \textbf{Size}$\downarrow$ & \textbf{PSNR}$\uparrow$ & \textbf{SSIM}$\uparrow$ & \textbf{LPIPS}$\downarrow$ & \textbf{Size}$\downarrow$ \\
\midrule

HAC-low & \colorbox{tabthird}{24.04} & 0.846 & 0.187 & \colorbox{tabsecond}{8.10} & 27.53 & 0.807 & 0.238 & 15.26 & 29.98 & 0.902 & 0.269 & \colorbox{tabthird}{4.35} \\
HAC-high & \colorbox{tabfirst}{24.40} & \colorbox{tabthird}{0.853} & \colorbox{tabthird}{0.177} & 11.24 & \colorbox{tabfirst}{27.77} & 0.811 & \colorbox{tabthird}{0.230} & 21.87 & \colorbox{tabthird}{30.34} & 0.906 & \colorbox{tabsecond}{0.258} & 6.35 \\
\addlinespace
IGS-low & 23.70 & 0.836 & 0.227 & 8.44 & 27.33 & 0.809 & 0.257 & 12.50 & 30.63 & 0.904 & 0.293 & 6.34 \\
IGS-high & \colorbox{tabsecond}{24.05} & 0.849 & 0.210 & 12.50 & \colorbox{tabthird}{27.62} & \colorbox{tabfirst}{0.819} & 0.247 & 25.40 & \colorbox{tabfirst}{32.33} & \colorbox{tabfirst}{0.924} & \colorbox{tabfirst}{0.253} & 7.74 \\
\addlinespace
SOG-w/o SH & 23.15 & 0.828 & 0.198 & 9.30 & 27.02 & 0.803 & \colorbox{tabsecond}{0.232} & 16.70 & \colorbox{tabsecond}{30.50} & \colorbox{tabthird}{0.908} & \colorbox{tabthird}{0.261} & 5.50 \\
SOG & 23.56 & 0.837 & 0.186 & 22.80 & \colorbox{tabsecond}{27.64} & \colorbox{tabsecond}{0.814} & \colorbox{tabfirst}{0.220} & 40.30 & 30.35 & \colorbox{tabsecond}{0.909} & \colorbox{tabsecond}{0.258} & 16.80 \\
\addlinespace
gsplat & 24.02 & \colorbox{tabfirst}{0.854} & \colorbox{tabfirst}{0.164} & 15.32 & 27.24 & \colorbox{tabthird}{0.808} & 0.231 & 14.82 & 29.87 & \colorbox{tabthird}{0.908} & 0.263 & 7.89 \\
\addlinespace
Ours ($2{\times}10^{-3}$) & 24.01 & \colorbox{tabthird}{0.853} & \colorbox{tabsecond}{0.165} & 10.66 & 27.20 & \colorbox{tabthird}{0.808} & \colorbox{tabsecond}{0.232} & \colorbox{tabthird}{10.98} & 29.77 & 0.905 & 0.270 & 4.76 \\
Ours ($6{\times}10^{-3}$) & 23.97 & \colorbox{tabsecond}{0.850} & 0.175 & \colorbox{tabthird}{8.41} & 26.94 & 0.801 & 0.243 & \colorbox{tabsecond}{9.04} & 29.16 & 0.893 & 0.296 & \colorbox{tabsecond}{3.68} \\
Ours ($1{\times}10^{-2}$) & 23.74 & 0.841 & 0.189 & \colorbox{tabfirst}{7.16} & 26.73 & 0.794 & 0.253 & \colorbox{tabfirst}{8.37} & 28.81 & 0.885 & 0.312 & \colorbox{tabfirst}{3.59} \\

\bottomrule
\end{tabular}
\label{tab:static_rd_table}
\end{table*}
\definecolor{tabfirst}{rgb}{1, 0.7, 0.7} 
\definecolor{tabsecond}{rgb}{1, 0.85, 0.7} 
\definecolor{tabthird}{rgb}{1, 1, 0.7} 

\begin{table}[t]
\centering

\small
\setlength{\tabcolsep}{4pt}

\caption{\textbf{The quantitative results of rate-distortion performance on dynamic scenes.} The \colorbox{tabfirst}{best}, \colorbox{tabsecond}{second best}, and \colorbox{tabthird}{third best} results are highlighted.}

\begin{tabular}{l|cccc}
\toprule
\multirow{2}{*}{\textbf{Method}} & \multicolumn{4}{c}{\textbf{Neural 3D Video}} \\
\cmidrule{2-5}
& \textbf{PSNR}$\uparrow$ & \textbf{SSIM}$\uparrow$ & \textbf{LPIPS}$\downarrow$ & \textbf{Rate}$\downarrow$ \\
\midrule

STG & \colorbox{tabsecond}{32.06} & \colorbox{tabsecond}{0.950} & \colorbox{tabfirst}{0.135} & 157.60 \\
\addlinespace
CompactSTG+PP & 31.69 & 0.945 & 0.148 & 12.32 \\
\addlinespace
Ours (GOF=50, high) & \colorbox{tabfirst}{32.17} & \colorbox{tabfirst}{0.951} & \colorbox{tabsecond}{0.138} & 13.50 \\
Ours (GOF=50, low) & \colorbox{tabthird}{31.98} & \colorbox{tabthird}{0.949} & 0.145 & \colorbox{tabthird}{12.15} \\
\addlinespace
Ours (GOF=300, high) & 31.86 & \colorbox{tabthird}{0.949} & \colorbox{tabthird}{0.139} & \colorbox{tabsecond}{5.06} \\
Ours (GOF=300, low) & 31.62 & 0.946 & 0.148 & \colorbox{tabfirst}{4.49} \\

\bottomrule
\end{tabular}

\label{tab:dynamic_rd_table}
\end{table}

\boldparagraph{Results}
Overall, our method achieves rate-distortion performance comparable to current baseline methods, shown in \figref{fig:static_rd_curve} and \tabref{tab:static_rd_table}. On Tanks\&Temples, we outperform the baselines in SSIM and LPIPS, while our PSNR-based performance ranks second to the top method, HAC, which compresses a more compact representation (Scaffold-GS~\cite{lu2024scaffold}), whereas we compress the original 3DGS. Compared to \gsplat compression, our method reduces the file size to 8 MB—with minimal quality loss. On MipNeRF360, our method performs slightly worse than HAC~\cite{chen2024hac} and IGS~\cite{wu2024IGS} in rate-distortion but requires significantly less storage while maintaining similar quality. Although our method currently doesn't reach the high-rate points of baselines, it is more competitive than PSNR in terms of SSIM and LPIPS. On DeepBlending, our method demonstrates comparable rate-distortion performance to other baseline approaches, achieving similar storage at comparable quality levels to HAC~\cite{chen2024hac} and SOG~\cite{morgenstern2023sogs}. Compared to our starting point, \gsplat~compression, our method achieves a 40\% reduction in storage with only a minimal drop in quality.

\begin{figure}[t]
    \centering
    \small
    \setlength{\tabcolsep}{1pt}
    \begin{tabular}{cccc}
        \includegraphics[width=0.24\columnwidth]{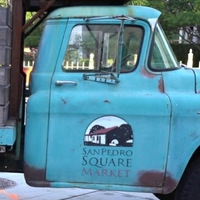} & 
        \includegraphics[width=0.24\columnwidth]{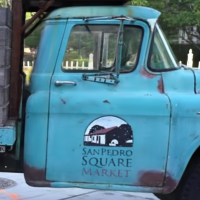} & 
        \includegraphics[width=0.24\columnwidth]{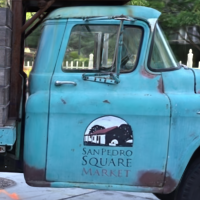} & 
        \includegraphics[width=0.24\columnwidth]{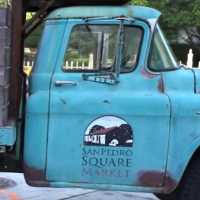} \\
        \includegraphics[width=0.24\columnwidth]{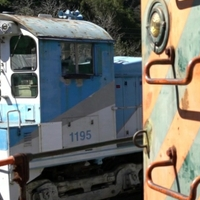} & 
        \includegraphics[width=0.24\columnwidth]{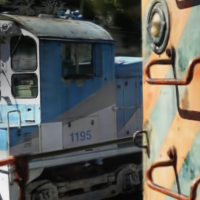} & 
        \includegraphics[width=0.24\columnwidth]{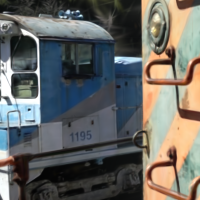} & 
        \includegraphics[width=0.24\columnwidth]{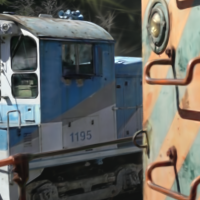} \\
        GT & HAC & \gsplat~comp. & Ours \\
    \end{tabular}
    \caption{\textbf{Qualitative comparisons} of static GS compression methods on the Tanks\&Temples dataset.}
    \label{fig:qualitative_static}
\end{figure}

Notably, \gsplat~compression serves as the starting point for our method and sets an upper limit for our performance. When \gsplat~compression quality approaches the highest baselines, our method achieves optimal rate-distortion performance, as SSIM and LPIPS on Tanks\&Temples demonstrated. A qualitative comparison in \figref{fig:qualitative_static} shows our method achieves a subjective quality similar to the baselines while using less storage.

\subsection{Dynamic GS Compression Comparison}

\boldparagraph{Datasets and baselines}
We conduct experiments on the Neural 3D Video dataset. The Neural 3D Video dataset contains six scenes, and our experiments are conducted on the first 300 frames of each scene. Our baseline method includes STG~\cite{li2024spacetimegaussian} and CompactSTG~\cite{lee2024C3DGS}. 

\boldparagraph{Settings}
In dynamic GS compression, the Group of Frames (GOF) is an important concept, referring to the number of consecutive frames grouped for joint reconstruction and compression, with different groups processed independently. STG and CompactSTG use a GOF value of 50. Our method supports two GOF configurations: 50 and 300. For instance, in a 300-frame sequence, a GOF of 50 means the sequence is divided into 6 segments, each containing 50 frames. The splats within each segment are trained and compressed together, while splats in different segments are processed independently.  Generally, a smaller GOF tends to yield higher reconstruction quality but at the cost of increased storage, as static content is redundantly represented across different GOFs. We achieve multiple rate points by adjusting the rate loss weight \(\lambda\), with values set to 0.02 and 0.01.

\begin{figure}[t]
    \centering
    \small
    \setlength{\tabcolsep}{1pt}
    
    \begin{tabular}{ccc}
        \includegraphics[width=0.32\columnwidth]{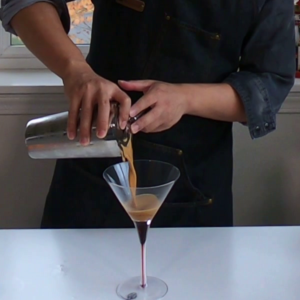} & 
        \includegraphics[width=0.32\columnwidth]{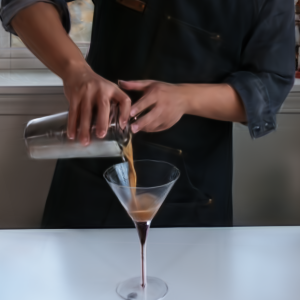} & 
        \includegraphics[width=0.32\columnwidth]{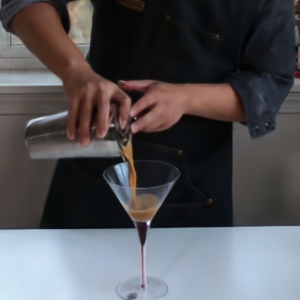} \\
        GT & CSTG &  Ours \\
    \end{tabular}
    \caption{\textbf{Qualitative comparisons} of dynamic GS compression methods on Neural 3D video dataset.}
    \label{fig:qualitative_dyn}
\end{figure}

\boldparagraph{Results}
We present the rate-distortion performance in \figref{fig:dyn_rd_curve} and \tabref{tab:dynamic_rd_table}.
In a fair comparison setting with GOF=50, our compositional method outperforms CompactSTG in rate-distortion efficiency. Compared to STG, our method achieves nearly a tenfold compression ratio without loss in quality. Increasing the GOF size further improves compression, though with some quality reduction. Specifically, with GOF=300, our method represents dynamic scenes at a 5 Mbps bitrate while maintaining quality similar to CompactSTG. 
We present qualitative results in \figref{fig:qualitative_dyn}. While our method shows clear metric advantages, it is subjectively similar to the baseline. This highlights the need for visual quality assessment methods for dynamic representations that better align with human visual perception.


\begin{table*}[htbp]
\vspace{-0.4cm}
\centering
\small
\setlength{\tabcolsep}{4pt}
\caption{\textbf{Ablations of our compression simulation options} on Tanks\&Temples dataset. {\color{green}\faCheck} refers to enable. {\color{red}\faTimes} refers to disable. We highlight the best results in bold font. } 
\begin{tabular}{l|>{\centering\arraybackslash}c>{\centering\arraybackslash}c|>{\centering\arraybackslash}c>{\centering\arraybackslash}c|>{\centering\arraybackslash}c>{\centering\arraybackslash}c|cccc}
\toprule
\multirow{2}{*}{\textbf{Methods}} & \multicolumn{2}{c|}{\textbf{Diff. Quantization}} & \multicolumn{2}{c|}{\textbf{Entropy Constraint}} & \multicolumn{2}{c|}{\textbf{Adaptive Masking}} & \multirow{2}{*}{\textbf{PSNR}} & \multirow{2}{*}{\textbf{SSIM}} & \multirow{2}{*}{\textbf{LPIPS}} & \multirow{2}{*}{\textbf{Mem.}} \\
\cmidrule(l{2pt}r{2pt}){2-3} \cmidrule(l{2pt}r{2pt}){4-5} \cmidrule(l{2pt}r{2pt}){6-7}
 & \textbf{Add Noise} & \textbf{STE-based} & \textbf{Factorized} & \textbf{Gaussian} & \textbf{Learnable} & \textbf{Gradient} & & & & \\
\midrule
Ours final                  & {\color{green}\faCheck} & {\color{red}\faTimes} & {\color{green}\faCheck} & {\color{red}\faTimes} & {\color{green}\faCheck} & {\color{red}\faTimes} & \textbf{23.74} & \textbf{0.840} & \textbf{0.189} & \textbf{7.10} \\
Opt to STE-based            & {\color{red}\faTimes} & {\color{green}\faCheck} & {\color{green}\faCheck} & {\color{red}\faTimes} & {\color{green}\faCheck} & {\color{red}\faTimes} & 23.47 & 0.819 & 0.221 & 7.73 \\
Opt to Gauss. model       & {\color{green}\faCheck} & {\color{red}\faTimes} & {\color{red}\faTimes} & {\color{green}\faCheck} & {\color{green}\faCheck} & {\color{red}\faTimes} & 23.68 & 0.834 & 0.195 & 8.01 \\
Opt to grad. thres.   & {\color{green}\faCheck} & {\color{red}\faTimes} & {\color{green}\faCheck} & {\color{red}\faTimes} & {\color{red}\faTimes} & {\color{green}\faCheck} & 23.62 & 0.839 & 0.192 & 7.28 \\
\bottomrule
\end{tabular}

\label{tab:ablation_comp_sim}
\end{table*}
\begin{table*}[htbp]
\centering
\small
\setlength{\tabcolsep}{4pt}
\caption{\textbf{Ablations of our post-training compression options} on Tanks\&Temples dataset. {\color{green}\faCheck} refers to enable. {\color{red}\faTimes} refers to disable. We highlight the best results in bold font. }
\begin{tabular}{l|c|cc|c|cc|cccc}
\toprule
\multirow{2}{*}{\textbf{Methods}} & \multirow{2}{*}{\textbf{Pruning}} & \multicolumn{2}{c|}{\textbf{Quantization}} & \multirow{2}{*}{\textbf{Mapping}} & \multicolumn{2}{c|}{\textbf{Coding Principles}} & \multirow{2}{*}{\textbf{PSNR}} & \multirow{2}{*}{\textbf{SSIM}} & \multirow{2}{*}{\textbf{LPIPS}} & \multirow{2}{*}{\textbf{Mem.}} \\
\cmidrule(l{2pt}r{2pt}){3-4} \cmidrule(l{2pt}r{2pt}){6-7}
& & \multicolumn{1}{c}{\textbf{8 bit}} & \multicolumn{1}{c|}{\textbf{6 bit}} & & \textbf{Learned Dist. Coding} & \textbf{Image Coding} & & & & \\
\midrule
Our final & {\color{green}\faCheck} & {\color{green}\faCheck} & {\color{red}\faTimes} & {\color{green}\faCheck} & {\color{green}\faCheck} & {\color{green}\faCheck} & \textbf{23.74} & \textbf{0.840} & \textbf{0.189} & 7.10 \\
Turn off pruning & {\color{red}\faTimes} & {\color{green}\faCheck} & {\color{red}\faTimes} & {\color{green}\faCheck} & {\color{green}\faCheck} & {\color{green}\faCheck} & \textbf{23.74} & \textbf{0.840} & \textbf{0.189} & 9.19 \\
Low bit-width & {\color{green}\faCheck} & {\color{red}\faTimes} & {\color{green}\faCheck} & {\color{green}\faCheck} & {\color{green}\faCheck} & {\color{green}\faCheck} & 23.54 & 0.834 & 0.195 & \textbf{6.10} \\
Only img. coding & {\color{green}\faCheck} & {\color{green}\faCheck} & {\color{red}\faTimes} & {\color{green}\faCheck} & {\color{red}\faTimes} & {\color{green}\faCheck} & \textbf{23.74} & \textbf{0.840} & \textbf{0.189} & 7.16 \\
\bottomrule
\end{tabular}

\label{tab:ablation_post_comp}
\end{table*}
\begin{table}[t]
    \small
    \centering
    \caption{\textbf{Analysis of the overall effectiveness of compression simulation and post-training compression.} ``w/o comp. sim." refers to training without compression simulation. ``w/o p.t. comp." refers to disabling the post-training compression.}
    \begin{tabular}{l|ccccc}
        \toprule
        \textbf{Method} & \textbf{PSNR} & \textbf{SSIM} & \textbf{LPIPS} & \textbf{\#Pts.} & \textbf{\#Mem.} \\
        \midrule
        w/o comp. sim. & 24.13 & 0.856 & 0.163 & 1 M. & 14 MB \\
        w/o p.t. comp. & 23.95 & 0.846 & 0.183 & 1 M. & 236 MB \\
        Ours final & 23.74 & 0.840 & 0.189 & 0.7 M. & 7 MB \\
        \bottomrule
    \end{tabular}

    \label{tab:overall_abl}
\end{table}
\begin{table}[t]
\centering
\small
\setlength{\tabcolsep}{6pt}
\caption{\textbf{Quantitative comparison of \gsplat~and feature-based splat methods on static scenes.} ``M." refers to Million.}
\begin{tabular}{l|c|cccc}
\toprule
\textbf{Scene} & \textbf{Method} & \textbf{PSNR} & \textbf{SSIM} & \textbf{LPIPS} & \textbf{\#Params} \\
\midrule
\multirow{2}{*}{Train}  & gsplat        & 22.38 & 0.832 & 0.186 & 59 M. \\
                        & feat. splat   & 22.34 & 0.823 & 0.197 & 17 M. \\
\midrule
\multirow{2}{*}{Truck}  & gsplat        & 25.97 & 0.891 & 0.125 & 59 M. \\
                        & feat. splat   & 25.64 & 0.882 & 0.135 & 17 M. \\
\bottomrule
\end{tabular}

\label{tab:static_rep}
\end{table}
\begin{table}[t]
\centering
\small
\setlength{\tabcolsep}{5.5pt}
\caption{\textbf{Quantitative comparison of different dynamic GS representation.} ``M." refers to Million. ``Poly" refers to Polynomial-based motion representation. ``Opa" refers to opacity. ``B.\&C." refers to Basis and Coefficients-based motion representation. }
\begin{tabular}{l|cccc}
\toprule
\textbf{Method} & \textbf{PSNR} & \textbf{SSIM} & \textbf{LPIPS} & \textbf{\#Params} \\
\midrule
\makecell[l]{Poly. + Varying Opa.} & 32.15 & 0.951 & 0.138 & 3.6 M. \\ 
\makecell[l]{B.\&C. +  Const. Opa.} & 30.42 & 0.934 & 0.147 & 11 M.\\ 
\bottomrule
\end{tabular}
\label{tab:dyn_rep}
\end{table}
\begin{table}
\small
\centering
\caption{\textbf{Memory consumption breakdown.}``Quat." refers to Quaternion. ``Opa." refers to Opacity. ``SH 0" refers to the zero-degree coefficients of SH. ``SH N" refers to the high-degree coefficients of SH. }
    \begin{tabular}{c|c|c}
    \toprule
    \textbf{Attributes} & \textbf{Compressed Size (KB)} & \textbf{Percentage} \\
    \midrule
    Mean    & 2585  & 40.6\%    \\
    Quat.  & 896   & 14.1\%    \\
    Scale   & 658   & 10.3\%    \\
    Opa.    & 277   & 4.3\%    \\
    SH 0    & 1033  & 16.2\%    \\
    SH N    & 923   & 14.5\%    \\
    \midrule
    Total   & 6372  & 100\% \\
    \bottomrule
    \end{tabular}
\label{tab:mem_breakdown}
\end{table}

\subsection{Analysis of Compression Simulation}
We demonstrate the effectiveness of training-time compression simulation in \tabref{tab:overall_abl}. Without it, memory usage reaches 14 MB—nearly double the final version—though quality improves slightly by 0.39 dB. This highlights its contribution to the final compression ratio. Next, we provide an in-depth analysis and discussion of the options for training-time compression simulation.

\boldparagraph{Compression Simulation Options}

\boldparagraph{Discussion on Differentiable Quantization}
As shown in the second row of \tabref{tab:ablation_comp_sim}, 
experimental results show that adding uniform noise, rather than using STE, for differentiable quantization yields better visual quality and compression performance.
In theory, the difference lies in the forward pass, as both share the same mechanism in the backward pass of the gradient.
The possible explanation is that uniform noise introduces stochastic perturbations, allowing post-quantization values to traverse a more continuous space, which facilitates parameter optimization toward solutions that achieve superior rate-distortion performance during training. In contrast, STE deterministically rounds values to the nearest quantized candidates, restricting post-quantization values to a discrete set and limiting the exploration of optimal solutions during training.

\begin{figure}[t]
    \centering
    \begin{tabular}{c}
    \includegraphics[width=0.95\linewidth]{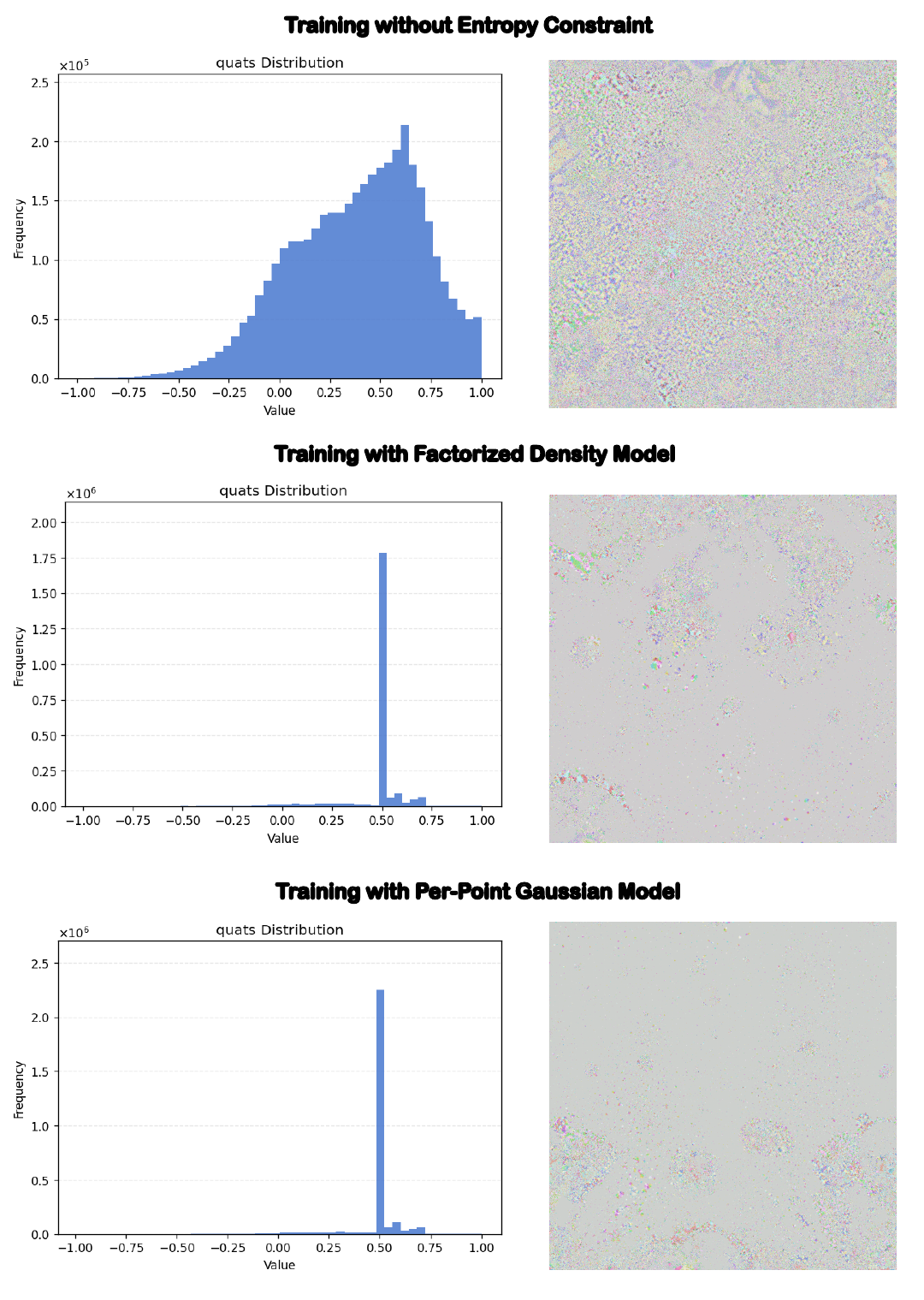}
    \end{tabular}
\caption{\textbf{Visualization on the effectiveness of entropy constraint.} As an example, we select the quaternion attribute from the attribute set for visualization. The left column visualizes the parameter distribution of quaternions, while the right column shows the quaternion map obtained after sorting.}
    \label{fig:entropy}
\end{figure}

\boldparagraph{Discussion on Entropy Constraint}
As shown in the third row of \tabref{tab:ablation_comp_sim}, we use the factorized density model instead of the per-point Gaussian model because it leads to less storage usage for attributes subject to the entropy constraint, resulting in nearly 1 MB of storage savings. 
In practice, we observed that adding or omitting entropy constraints during training has a significant impact on the final parameter distribution, which directly affects compression performance.
We visualize the parameter distributions to examine the effect of the entropy constraint. As shown in \figref{fig:entropy}, the distribution of Gaussian Splat parameters trained with entropy constraints is much more compact than those without entropy constraints, leading to lower entropy in the final parameters, which benefits post-training compression. Next, we visualize the 2D maps of the elements to show the impact of the entropy constraint. \figref{fig:entropy} shows that the entropy constraint smooths the sorted quaternion map, reducing high-frequency components and making it easier for 2D image codecs to compress.

Interestingly, both the factorized density model and the per-point Gaussian distribution model yield compact parameter distributions when used in entropy constraints.
The factorized density model inherently constrains all parameters to share a complex distribution parameterized by an MLP. It is reasonable that the Gaussian Splat parameters and the learned distribution converge to a compact distribution via a rate-distortion loss.
The per-point Gaussian model also converges to a compact distribution, though this may seem counterintuitive. One possible explanation is that the rate loss not only pushes the predicted per-point mean and variance from the learned hash table closer to the Gaussian Splat parameters but also influences the Gaussian Splat parameters to shift toward values that simplify mean and variance estimation from the hash table. This bidirectional influence may drive the Gaussian Splat parameters to gradually cluster around certain values.

Experimentally, the factorized density model achieves better rate-distortion performance. This advantage likely comes from its ability to model more complex distributions. The underlying mechanism is worth further investigation.

\boldparagraph{Discussion on Adaptive Masking}
For adaptive masking, experiments demonstrate that a learnable mask slightly outperforms gradient thresholding, shown in the last row of \tabref{tab:ablation_comp_sim}. The mask ratio is a key hyperparameter.
Gradient thresholding makes hard decisions. During training, a splat is marked as holding view-dependent color if its SH coefficients exceed a gradient threshold in any backward pass. When the mask ratio is reached, no further splats are included. Thus, it often leads to suboptimal results.
In contrast, the learnable mask optimizes in a continuous space and is binarized when applied. The mask ratio is incorporated through KL loss, allowing it to adapt based on content. This flexibility leads to better experimental results.

\boldparagraph{Findings}
In short, our analysis of compression simulation techniques provides several key insights:
\begin{itemize}
    \item \textit{Differentiable Quantization:} Uniform noise outperforms STE by allowing continuous exploration of optimal solutions during training, achieving better rate-distortion performance.
    \item \textit{Entropy Constraint:} Adding an entropy constraint during training compacts parameter distributions, enhancing compression performance.
    \item \textit{Adaptive Masking:} Learnable masks offer superior rate-distortion outcomes due to their flexibility.
\end{itemize}

\subsection{Analysis of Post-Training Compression}
We also show the overall effectiveness of post-training compression in \tabref{tab:overall_abl}. As shown in the second row of \tabref{tab:overall_abl}, when only compression simulation is used without post-training compression, memory usage exceeds 200 MB, highlighting the impact of post-training compression.
Next, we provide discussion and analysis on the design choices for post-training compression.

\boldparagraph{Discussion on Pruning}
In Static GSCodec, post-training pruning proves highly effective, reflecting the presence of many splats with extremely low opacity after training. As shown in the second row of \tabref{tab:ablation_post_comp}, applying pruning significantly reduces the final file size without a quality drop. The underlying reason lies in the entropy constraint applied to opacity during training. The rate-distortion loss drives many splats toward very low opacity values, especially if they do not contribute much to improving rendering quality.

\boldparagraph{Discussion on Low Bit-Width Quantization}
When using a 6-bit width for scalar uniform quantization, there is a noticeable quality drop, though it reduces storage by nearly 1MB, as shown in the third row of \tabref{tab:ablation_post_comp}. To maintain optimal quality, we opt for to 8-bit width for scalar uniform quantization. 
Additionally, designing attribute-specific non-uniform quantization schemes tailored to the nature of their distributions holds promise for future exploration and could further enhance the compression ratio.

\boldparagraph{Discussion on Lossless Coding}
For lossless coding methods after quantization, we provide both learned-distribution-based entropy coding and lossless image coding via PNG compression. Using only image coding shows only a slight disadvantage in the last row of \tabref{tab:ablation_post_comp}. Although learned-distribution-based entropy coding theoretically offers a higher compression ratio, the gap between the learned and actual distributions likely limits its effectiveness, resulting in marginal compression gains.

\boldparagraph{Findings}
In short, our analysis of post-training compression techniques provides several key insights:
\begin{itemize}
    \item \textit{Pruning:} Post-training pruning, combined with a training-time entropy constraint on opacity, removes numerous splats that do not contribute to rendering quality.
    \item \textit{Quantization Bit-Width:} An 8-bit depth is a suitable choice for uniform scalar quantization.
    \item \textit{Lossless Coding:} 
    Entropy coding based on learned distributions offers modest gains in compression performance but still has large room for improvement.
\end{itemize}

\subsection{Ablations on Representation Choices}

\boldparagraph{Static Representations}
We compare two static representations, \gsplat~and feature-based splat, focusing on quality metrics and parameter counts in \tabref{tab:static_rep}. Both methods use MCMC~\cite{kheradmand20243dgsmcmc} as the sampling strategy. \gsplat~outperforms feature-based splat in quality, especially in reflective scenes like the Truck. The feature-based splat, however, achieves acceptable rendering quality with fewer parameters. We select \gsplat~as the representative static representation to ensure the best possible quality.

\boldparagraph{Dynamic Representations}
We compare two dynamic representations: one based on polynomial motion with time-varying opacity, and the other using basis and coefficients to model motion with time-invariant opacity in \tabref{tab:dyn_rep}. The first approach is similar to STG~\cite{li2024spacetimegaussian}, while the second resembles DynMF~\cite{kratimenos2024dynmf} and Shape of Motion~\cite{wang2024shapeofmotion}.
Focusing on representation, we report only quality metrics and parameter counts. Our experiments show the polynomial motion representation with time-varying opacity provides better quality and requires fewer parameters. As a result, we select this approach as our dynamic representation.

\subsection{Analysis of Memory Consumption}
\boldparagraph{Results}
We report the memory consumption of different components from the Truck scene as an example, shown in \tabref{tab:mem_breakdown}. The position information takes up the majority of the storage, accounting for over 40\% of the total usage. 
Thanks to adaptive masking and efficient VQ, high-degree SH coefficients named ``SH N", which originally occupied a significant portion of the data, now takes up less than 15\% after compression. 

\boldparagraph{Discussion and Findings}
An analysis of memory consumption reveals that position data accounts for the majority of storage requirements. The key challenge ahead is finding effective methods to compress position information while maintaining high compression ratios for other attributes.
Additionally, our experiments indicate that the high-degree SH coefficients in the vanilla 3DGS representation contain significant redundancy.

\section{Conclusion and future work}
We propose GSCodec Studio, a modular framework for Gaussian Splat compression that integrates high-quality static and dynamic GS representations with efficient compression methods. Serving as an experimental platform, it aims to advance the field.

This framework includes four key stages in the GS compression pipeline: training data preparation, training with compression simulation, post-training compression, and viewing. For each stage, we have developed core components for GS compression, offering at least two solutions for research exploration. These components cover static and dynamic representations, training-time compression, and post-training compression techniques. The modular design enables versatile prototyping and experimentation.

Building on modular core components, we conducted extensive experiments and delivered promising static and dynamic GS compression solutions, namely Static GSCodec and Dynamic GSCodec, which incorporate best practices. Notably, the Dynamic GSCodec achieves state-of-the-art rate-distortion performance in dynamic GS compression. Our ablation studies reveal that entropy constraints during training effectively improve compressibility, while efficient compression of positions in Gaussian Splats remains a key challenge.

In the future, we plan to expand the framework with more efficient representations and compression techniques to further support advancements in GS compression research.

{
    \bibliographystyle{IEEEtran}
    \bibliography{bibliography, bibliography_long, bibliography_custom}

\begin{thebibliography}{10}
\providecommand{\url}[1]{#1}
\csname url@samestyle\endcsname
\providecommand{\newblock}{\relax}
\providecommand{\bibinfo}[2]{#2}
\providecommand{\BIBentrySTDinterwordspacing}{\spaceskip=0pt\relax}
\providecommand{\BIBentryALTinterwordstretchfactor}{4}
\providecommand{\BIBentryALTinterwordspacing}{\spaceskip=\fontdimen2\font plus
\BIBentryALTinterwordstretchfactor\fontdimen3\font minus
  \fontdimen4\font\relax}
\providecommand{\BIBforeignlanguage}[2]{{%
\expandafter\ifx\csname l@#1\endcsname\relax
\typeout{** WARNING: IEEEtran.bst: No hyphenation pattern has been}%
\typeout{** loaded for the language `#1'. Using the pattern for}%
\typeout{** the default language instead.}%
\else
\language=\csname l@#1\endcsname
\fi
#2}}
\providecommand{\BIBdecl}{\relax}
\BIBdecl

\bibitem{kerbl3Dgaussians}
\BIBentryALTinterwordspacing
B.~Kerbl, G.~Kopanas, T.~Leimk{\"u}hler, and G.~Drettakis, ``3d gaussian
  splatting for real-time radiance field rendering,'' \emph{ACM Trans. on
  Graphics}, vol.~42, no.~4, July 2023. [Online]. Available:
  \url{https://repo-sam.inria.fr/fungraph/3d-gaussian-splatting/}
\BIBentrySTDinterwordspacing

\bibitem{luiten2023dynamic3DG}
J.~Luiten, G.~Kopanas, B.~Leibe, and D.~Ramanan, ``Dynamic 3d gaussians:
  Tracking by persistent dynamic view synthesis,'' in \emph{3DV}, 2024.

\bibitem{Wu20244dgaussiansplatting}
G.~Wu, T.~Yi, J.~Fang, L.~Xie, X.~Zhang, W.~Wei, W.~Liu, Q.~Tian, and X.~Wang,
  ``4d gaussian splatting for real-time dynamic scene rendering,'' in
  \emph{Proc. IEEE Conf. on Computer Vision and Pattern Recognition (CVPR)},
  2024.

\bibitem{yang2023deformable3dgs}
Z.~Yang, X.~Gao, W.~Zhou, S.~Jiao, Y.~Zhang, and X.~Jin, ``Deformable 3d
  gaussians for high-fidelity monocular dynamic scene reconstruction,''
  \emph{Proc. IEEE Conf. on Computer Vision and Pattern Recognition (CVPR)},
  2024.

\bibitem{yang2023gs4d}
Z.~Yang, H.~Yang, Z.~Pan, and L.~Zhang, ``Real-time photorealistic dynamic
  scene representation and rendering with 4d gaussian splatting,'' in
  \emph{Proc. of the International Conf. on Learning Representations (ICLR)},
  2024.

\bibitem{li2024spacetimegaussian}
Z.~Li, Z.~Chen, Z.~Li, and Y.~Xu, ``Spacetime gaussian feature splatting for
  real-time dynamic view synthesis,'' in \emph{Proc. IEEE Conf. on Computer
  Vision and Pattern Recognition (CVPR)}, 2024.

\bibitem{kratimenos2024dynmf}
A.~Kratimenos, J.~Lei, and K.~Daniilidis, ``Dynmf: Neural motion factorization
  for real-time dynamic view synthesis with 3d gaussian splatting,''
  \emph{arXiV}, 2023.

\bibitem{bao20253dgssurvey}
Y.~Bao, T.~Ding, J.~Huo, Y.~Liu, Y.~Li, W.~Li, Y.~Gao, and J.~Luo, ``3d
  gaussian splatting: Survey, technologies, challenges, and opportunities,''
  \emph{IEEE Transactions on Circuits and Systems for Video Technology}, 2025.

\bibitem{charatan23pixelsplat}
D.~Charatan, S.~Li, A.~Tagliasacchi, and V.~Sitzmann, ``pixelsplat: 3d gaussian
  splats from image pairs for scalable generalizable 3d reconstruction,'' in
  \emph{Proc. IEEE Conf. on Computer Vision and Pattern Recognition (CVPR)},
  2024.

\bibitem{Yu2024MipSplatting}
Z.~Yu, A.~Chen, B.~Huang, T.~Sattler, and A.~Geiger, ``Mip-splatting:
  Alias-free 3d gaussian splatting,'' in \emph{Proc. IEEE Conf. on Computer
  Vision and Pattern Recognition (CVPR)}, June 2024.

\bibitem{kheradmand20243dgsmcmc}
S.~Kheradmand, D.~Rebain, G.~Sharma, W.~Sun, J.~Tseng, H.~Isack, A.~Kar,
  A.~Tagliasacchi, and K.~M. Yi, ``3d gaussian splatting as markov chain monte
  carlo,'' \emph{arXiv preprint arXiv:2404.09591}, 2024.

\bibitem{chen2024mvsplat}
Y.~Chen, H.~Xu, C.~Zheng, B.~Zhuang, M.~Pollefeys, A.~Geiger, T.-J. Cham, and
  J.~Cai, ``Mvsplat: Efficient 3d gaussian splatting from sparse multi-view
  images,'' \emph{Proc. of the European Conf. on Computer Vision (ECCV)}, 2024.

\bibitem{Huang20242DGS}
B.~Huang, Z.~Yu, A.~Chen, A.~Geiger, and S.~Gao, ``2d gaussian splatting for
  geometrically accurate radiance fields,'' in \emph{ACM Trans. on Graphics},
  2024.

\bibitem{hu2024gaussianavatar}
L.~Hu, H.~Zhang, Y.~Zhang, B.~Zhou, B.~Liu, S.~Zhang, and L.~Nie,
  ``Gaussianavatar: Towards realistic human avatar modeling from a single video
  via animatable 3d gaussians,'' in \emph{Proc. IEEE Conf. on Computer Vision
  and Pattern Recognition (CVPR)}, 2024.

\bibitem{qian2024gaussianavatars}
S.~Qian, T.~Kirschstein, L.~Schoneveld, D.~Davoli, S.~Giebenhain, and
  M.~Nie{\ss}ner, ``Gaussianavatars: Photorealistic head avatars with rigged 3d
  gaussians,'' \emph{Proc. IEEE Conf. on Computer Vision and Pattern
  Recognition (CVPR)}, 2024.

\bibitem{saito2024relightable}
S.~Saito, G.~Schwartz, T.~Simon, J.~Li, and G.~Nam, ``Relightable gaussian
  codec avatars,'' in \emph{Proc. IEEE Conf. on Computer Vision and Pattern
  Recognition (CVPR)}, 2024, pp. 130--141.

\bibitem{zheng2024gpsgaussian}
S.~Zheng, B.~Zhou, R.~Shao, B.~Liu, S.~Zhang, L.~Nie, and Y.~Liu,
  ``Gps-gaussian: Generalizable pixel-wise 3d gaussian splatting for real-time
  human novel view synthesis,'' in \emph{Proceedings of the IEEE/CVF Conference
  on Computer Vision and Pattern Recognition}, 2024, pp. 19\,680--19\,690.

\bibitem{zhang2025humanrefgs}
J.~Zhang, X.~Li, H.~Zhong, Q.~Zhang, Y.~Cao, Y.~Shan, and J.~Liao,
  ``Humanref-gs: Image-to-3d human generation with reference-guided diffusion
  and 3d gaussian splatting,'' \emph{IEEE Transactions on Circuits and Systems
  for Video Technology}, 2025.

\bibitem{tang2024dreamgaussian}
J.~Tang, J.~Ren, H.~Zhou, Z.~Liu, and G.~Zeng, ``Dreamgaussian: Generative
  gaussian splatting for efficient 3d content creation,'' \emph{Proc. of the
  International Conf. on Learning Representations (ICLR)}, 2024.

\bibitem{xu2024grm}
Y.~Xu, Z.~Shi, W.~Yifan, H.~Chen, C.~Yang, S.~Peng, Y.~Shen, and G.~Wetzstein,
  ``Grm: Large gaussian reconstruction model for efficient 3d reconstruction
  and generation,'' \emph{arXiv preprint arXiv:2403.14621}, 2024.

\bibitem{yan2024GSSLAM}
C.~Yan, D.~Qu, D.~Xu, B.~Zhao, Z.~Wang, D.~Wang, and X.~Li, ``Gs-slam: Dense
  visual slam with 3d gaussian splatting,'' in \emph{Proc. IEEE Conf. on
  Computer Vision and Pattern Recognition (CVPR)}, 2024.

\bibitem{keetha2024splatam}
N.~Keetha, J.~Karhade, K.~M. Jatavallabhula, G.~Yang, S.~Scherer, D.~Ramanan,
  and J.~Luiten, ``Splatam: Splat, track \& map 3d gaussians for dense rgb-d
  slam,'' in \emph{Proc. IEEE Conf. on Computer Vision and Pattern Recognition
  (CVPR)}, 2024.

\bibitem{tu2024telealoha}
H.~Tu, R.~Shao, X.~Dong, S.~Zheng, H.~Zhang, L.~Chen, M.~Wang, W.~Li, S.~Ma,
  S.~Zhang \emph{et~al.}, ``Tele-aloha: A low-budget and high-authenticity
  telepresence system using sparse rgb cameras,'' \emph{ACM Trans. on
  Graphics}, 2024.

\bibitem{cui2025streetsurfgs}
X.~Cui, W.~Ye, Y.~Wang, G.~Zhang, W.~Zhou, T.~He, and H.~Li, ``Streetsurfgs:
  Scalable urban street surface reconstruction with planar-based gaussian
  splatting,'' \emph{IEEE Transactions on Circuits and Systems for Video
  Technology}, 2025.

\bibitem{xie2024physgaussian}
T.~Xie, Z.~Zong, Y.~Qiu, X.~Li, Y.~Feng, Y.~Yang, and C.~Jiang, ``Physgaussian:
  Physics-integrated 3d gaussians for generative dynamics,'' \emph{Proc. IEEE
  Conf. on Computer Vision and Pattern Recognition (CVPR)}, 2024.

\bibitem{jiang2024vrgs}
Y.~Jiang, C.~Yu, T.~Xie, X.~Li, Y.~Feng, H.~Wang, M.~Li, H.~Lau, F.~Gao,
  Y.~Yang, and C.~Jiang, ``Vr-gs: A physical dynamics-aware interactive
  gaussian splatting system in virtual reality,'' \emph{ACM Trans. on
  Graphics}, 2024.

\bibitem{abou2024physically}
J.~Abou-Chakra, K.~Rana, F.~Dayoub, and N.~S{\"u}nderhauf, ``Physically
  embodied gaussian splatting: A realtime correctable world model for
  robotics,'' in \emph{8th Annual Conference on Robot Learning}, 2024.

\bibitem{luiten2023dynamic3dgs}
J.~Luiten, G.~Kopanas, B.~Leibe, and D.~Ramanan, ``Dynamic 3d gaussians:
  Tracking by persistent dynamic view synthesis,'' in \emph{Proc. of the
  International Conf. on 3D Vision (3DV)}, 2024.

\bibitem{sun20243dgstream}
J.~Sun, H.~Jiao, G.~Li, Z.~Zhang, L.~Zhao, and W.~Xing, ``3dgstream: On-the-fly
  training of 3d gaussians for efficient streaming of photo-realistic
  free-viewpoint videos,'' in \emph{Proc. IEEE Conf. on Computer Vision and
  Pattern Recognition (CVPR)}, 2024.

\bibitem{guo2024motion3dgs}
Z.~Guo, W.~Zhou, L.~Li, M.~Wang, and H.~Li, ``Motion-aware 3d gaussian
  splatting for efficient dynamic scene reconstruction,'' \emph{IEEE
  Transactions on Circuits and Systems for Video Technology}, 2024.

\bibitem{li2025frpgs}
W.~Li, X.~Pan, J.~Lin, P.~Lu, D.~Feng, and W.~Shi, ``Frpgs: Fast, robust, and
  photorealistic monocular dynamic scene reconstruction with deformable 3d
  gaussians,'' \emph{IEEE Transactions on Circuits and Systems for Video
  Technology}, 2025.

\bibitem{huang2024scgs}
Y.-H. Huang, Y.-T. Sun, Z.~Yang, X.~Lyu, Y.-P. Cao, and X.~Qi, ``Sc-gs:
  Sparse-controlled gaussian splatting for editable dynamic scenes,'' in
  \emph{Proc. IEEE Conf. on Computer Vision and Pattern Recognition (CVPR)},
  2024.

\bibitem{wang2024shapeofmotion}
Q.~Wang, V.~Ye, H.~Gao, J.~Austin, Z.~Li, and A.~Kanazawa, ``Shape of motion:
  4d reconstruction from a single video,'' 2024.

\bibitem{duan20244drotorgs}
Y.~Duan, F.~Wei, Q.~Dai, Y.~He, W.~Chen, and B.~Chen, ``4d-rotor gaussian
  splatting: Towards efficient novel view synthesis for dynamic scenes,'' in
  \emph{ACM Trans. on Graphics}, 2024, pp. 1--11.

\bibitem{lu2024scaffold}
T.~Lu, M.~Yu, L.~Xu, Y.~Xiangli, L.~Wang, D.~Lin, and B.~Dai, ``Scaffold-gs:
  Structured 3d gaussians for view-adaptive rendering,'' in \emph{Proc. IEEE
  Conf. on Computer Vision and Pattern Recognition (CVPR)}, 2024.

\bibitem{ververas2024sags}
E.~Ververas, R.~A. Potamias, J.~Song, J.~Deng, and S.~Zafeiriou, ``Sags:
  Structure-aware 3d gaussian splatting,'' \emph{arXiv preprint
  arXiv:2404.19149}, 2024.

\bibitem{papantonakis2024reducing}
P.~Papantonakis, G.~Kopanas, B.~Kerbl, A.~Lanvin, and G.~Drettakis, ``Reducing
  the memory footprint of 3d gaussian splatting,'' \emph{Proceedings of the ACM
  on Computer Graphics and Interactive Techniques}, vol.~7, no.~1, pp. 1--17,
  2024.

\bibitem{fan2023lightgaussian}
Z.~Fan, K.~Wang, K.~Wen, Z.~Zhu, D.~Xu, and Z.~Wang, ``Lightgaussian: Unbounded
  3d gaussian compression with 15x reduction and 200+ fps,'' \emph{arXiv
  preprint arXiv:2311.17245}, 2023.

\bibitem{lee2024Compact3DGaussian}
J.~C. Lee, D.~Rho, X.~Sun, J.~H. Ko, and E.~Park, ``Compact 3d gaussian
  representation for radiance field,'' in \emph{Proc. IEEE Conf. on Computer
  Vision and Pattern Recognition (CVPR)}, 2024, pp. 21\,719--21\,728.

\bibitem{lee2024C3DGS}
J.~C. Lee, D.~Rho, X.~Sun, J.~H. Ko, and E.~B. Park, ``Compact 3d gaussian
  splatting for static and dynamic radiance fields,'' \emph{arXiv preprint
  arXiv:2408.03822}, 2024.

\bibitem{fang2024minisplatting}
G.~Fang and B.~Wang, ``Mini-splatting: Representing scenes with a constrained
  number of gaussians,'' \emph{arXiv preprint arXiv:2403.14166}, 2024.

\bibitem{navaneet2023compact3d}
K.~Navaneet, K.~P. Meibodi, S.~A. Koohpayegani, and H.~Pirsiavash, ``Compgs:
  Smaller and faster gaussian splatting with vector quantization,''
  \emph{ECCV}, 2024.

\bibitem{niedermayr2024compressed3dgs}
S.~Niedermayr, J.~Stumpfegger, and R.~Westermann, ``Compressed 3d gaussian
  splatting for accelerated novel view synthesis,'' in \emph{Proc. IEEE Conf.
  on Computer Vision and Pattern Recognition (CVPR)}, 2024.

\bibitem{chen2024hac}
Y.~Chen, Q.~Wu, J.~Cai, M.~Harandi, and W.~Lin, ``Hac: Hash-grid assisted
  context for 3d gaussian splatting compression,'' \emph{Proc. of the European
  Conf. on Computer Vision (ECCV)}, 2024.

\bibitem{morgenstern2023sogs}
W.~Morgenstern, F.~Barthel, A.~Hilsmann, and P.~Eisert, ``Compact 3d scene
  representation via self-organizing gaussian grids,'' \emph{Proc. of the
  European Conf. on Computer Vision (ECCV)}, 2024.

\bibitem{xie2025mesongs}
S.~Xie, W.~Zhang, C.~Tang, Y.~Bai, R.~Lu, S.~Ge, and Z.~Wang, ``Mesongs:
  Post-training compression of 3d gaussians via efficient attribute
  transformation,'' in \emph{European Conference on Computer Vision}.\hskip 1em
  plus 0.5em minus 0.4em\relax Springer, 2025, pp. 434--452.

\bibitem{wang2024contextgs}
Y.~Wang, Z.~Li, L.~Guo, W.~Yang, A.~C. Kot, and B.~Wen, ``Contextgs: Compact 3d
  gaussian splatting with anchor level context model,'' \emph{arXiv preprint
  arXiv:2405.20721}, 2024.

\bibitem{wang2024rdo3dgs}
H.~Wang, H.~Zhu, T.~He, R.~Feng, J.~Deng, J.~Bian, and Z.~Chen, ``End-to-end
  rate-distortion optimized 3d gaussian representation,'' \emph{Proc. of the
  European Conf. on Computer Vision (ECCV)}, 2024.

\bibitem{liu2024compgs}
X.~Liu, X.~Wu, P.~Zhang, S.~Wang, Z.~Li, and S.~Kwong, ``Compgs: Efficient 3d
  scene representation via compressed gaussian splatting,'' in
  \emph{Proceedings of the 32nd ACM International Conference on Multimedia},
  2024, pp. 2936--2944.

\bibitem{Jiang_2024_CVPR}
Y.~Jiang, Z.~Shen, P.~Wang, Z.~Su, Y.~Hong, Y.~Zhang, J.~Yu, and L.~Xu,
  ``Hifi4g: High-fidelity human performance rendering via compact gaussian
  splatting,'' in \emph{Proceedings of the IEEE/CVF Conference on Computer
  Vision and Pattern Recognition (CVPR)}, June 2024, pp. 19\,734--19\,745.

\bibitem{jiang2024robust}
\BIBentryALTinterwordspacing
Y.~Jiang, Z.~Shen, Y.~Hong, C.~Guo, Y.~Wu, Y.~Zhang, J.~Yu, and L.~Xu, ``Robust
  dual gaussian splatting for immersive human-centric volumetric videos,''
  \emph{ACM Transactions on Graphics (TOG)}, vol.~43, no.~6, p.~15, Dec 2024.
  [Online]. Available: \url{https://doi.org/10.1145/3687926}
\BIBentrySTDinterwordspacing

\bibitem{wang2024v3}
\BIBentryALTinterwordspacing
P.~Wang, Z.~Zhang, L.~Wang, K.~Yao, S.~Xie, J.~Yu, M.~Wu, and L.~Xu, ``V3:
  Viewing volumetric videos on mobiles via streamable 2d dynamic gaussians,''
  \emph{ACM Transactions on Graphics (TOG)}, vol.~43, no.~6, p.~13, Dec 2024.
  [Online]. Available: \url{https://doi.org/10.1145/3687935}
\BIBentrySTDinterwordspacing

\bibitem{bae2024EmbedD3dgs}
J.~Bae, S.~Kim, Y.~Yun, H.~Lee, G.~Bang, and Y.~Uh, ``Per-gaussian
  embedding-based deformation for deformable 3d gaussian splatting,''
  \emph{arXiv preprint arXiv:2404.03613}, 2024.

\bibitem{zhang2024mega}
X.~Zhang, Z.~Liu, Y.~Zhang, X.~Ge, D.~He, T.~Xu, Y.~Wang, Z.~Lin, S.~Yan, and
  J.~Zhang, ``Mega: Memory-efficient 4d gaussian splatting for dynamic
  scenes,'' \emph{arXiv preprint arXiv:2410.13613}, 2024.

\bibitem{kwak2025modec}
S.~Kwak, J.~Kim, J.~Y. Jeong, W.-S. Cheong, J.~Oh, and M.~Kim, ``Modec-gs:
  Global-to-local motion decomposition and temporal interval adjustment for
  compact dynamic 3d gaussian splatting,'' \emph{arXiv preprint
  arXiv:2501.03714}, 2025.

\bibitem{girish2024queen}
S.~Girish, T.~Li, A.~Mazumdar, A.~Shrivastava, S.~De~Mello \emph{et~al.},
  ``Queen: Quantized efficient encoding of dynamic gaussians for streaming
  free-viewpoint videos,'' \emph{Advances in Neural Information Processing
  Systems (NeurIPS)}, vol.~37, pp. 43\,435--43\,467, 2024.

\bibitem{zhang2025evolvinggs}
C.~Zhang, Y.~Zhou, S.~Wang, W.~Li, D.~Wang, Y.~Xu, and S.~Jiao, ``Evolvinggs:
  High-fidelity streamable volumetric video via evolving 3d gaussian
  representation,'' \emph{arXiv preprint arXiv:2503.05162}, 2025.

\bibitem{ye2024gsplatopensourcelibrarygaussian}
\BIBentryALTinterwordspacing
V.~Ye, R.~Li, J.~Kerr, M.~Turkulainen, B.~Yi, Z.~Pan, O.~Seiskari, J.~Ye,
  J.~Hu, M.~Tancik, and A.~Kanazawa, ``gsplat: An open-source library for
  {Gaussian} splatting,'' \emph{arXiv preprint arXiv:2409.06765}, 2024.
  [Online]. Available: \url{https://arxiv.org/abs/2409.06765}
\BIBentrySTDinterwordspacing

\bibitem{ye2024gaustudio}
C.~Ye, Y.~Nie, J.~Chang, Y.~Chen, Y.~Zhi, and X.~Han, ``Gaustudio: A modular
  framework for 3d gaussian splatting and beyond,'' \emph{arXiv preprint
  arXiv:2403.19632}, 2024.

\bibitem{xu2023easyvolcap}
Z.~Xu, T.~Xie, S.~Peng, H.~Lin, Q.~Shuai, Z.~Yu, G.~He, J.~Sun, H.~Bao, and
  X.~Zhou, ``Easyvolcap: Accelerating neural volumetric video research,'' in
  \emph{SIGGRAPH Asia 2023 Technical Communications}, 2023, pp. 1--4.

\bibitem{nerfstudio}
M.~Tancik, E.~Weber, E.~Ng, R.~Li, B.~Yi, J.~Kerr, T.~Wang, A.~Kristoffersen,
  J.~Austin, K.~Salahi, A.~Ahuja, D.~McAllister, and A.~Kanazawa, ``Nerfstudio:
  A modular framework for neural radiance field development,'' \emph{arXiv
  preprint arXiv:2302.04264}, 2023.

\bibitem{chen2022mobilenerf}
Z.~Chen, T.~Funkhouser, P.~Hedman, and A.~Tagliasacchi, ``Mobilenerf:
  Exploiting the polygon rasterization pipeline for efficient neural field
  rendering on mobile architectures,'' \emph{arXiv.org}, 2022.

\bibitem{nagel2021nnquantization}
M.~Nagel, M.~Fournarakis, R.~A. Amjad, Y.~Bondarenko, M.~Van~Baalen, and
  T.~Blankevoort, ``A white paper on neural network quantization,'' \emph{arXiv
  preprint arXiv:2106.08295}, 2021.

\bibitem{balle2018hyperprior}
J.~Ball{\'e}, D.~Minnen, S.~Singh, S.~J. Hwang, and N.~Johnston, ``Variational
  image compression with a scale hyperprior,'' in \emph{Proc. of the
  International Conf. on Learning Representations (ICLR)}, 2018.

\bibitem{begaint2020compressai}
J.~B{\'e}gaint, F.~Racap{\'e}, S.~Feltman, and A.~Pushparaja, ``Compressai: a
  pytorch library and evaluation platform for end-to-end compression
  research,'' \emph{arXiv preprint arXiv:2011.03029}, 2020.

\bibitem{ascenso2023jpegai}
J.~Ascenso, E.~Alshina, and T.~Ebrahimi, ``The jpeg ai standard: Providing
  efficient human and machine visual data consumption,'' \emph{IEEE
  Multimedia}, vol.~30, no.~1, pp. 100--111, 2023.

\bibitem{making_GS_more_smaller}
\BIBentryALTinterwordspacing
A.~Pranckevičius. (2023) Making gaussian splats more smaller. [Online].
  Available:
  \url{https://aras-p.info/blog/2023/09/27/Making-Gaussian-Splats-more-smaller/}
\BIBentrySTDinterwordspacing

\bibitem{torchpq}
\BIBentryALTinterwordspacing
DeMoriarty. (2022) Torchpq. [Online]. Available:
  \url{https://github.com/DeMoriarty/TorchPQ}
\BIBentrySTDinterwordspacing

\bibitem{wu2024IGS}
M.~Wu and T.~Tuytelaars, ``Implicit gaussian splatting with efficient
  multi-level tri-plane representation,'' \emph{arXiv preprint
  arXiv:2408.10041}, 2024.

\bibitem{bamler2022constriction}
R.~Bamler, ``Understanding entropy coding with asymmetric numeral systems
  (ans): a statistician's perspective,'' \emph{arXiv preprint
  arXiv:2201.01741}, 2022.

\bibitem{duda2009ans}
J.~Duda, ``Asymmetric numeral systems,'' \emph{arXiv preprint arXiv:0902.0271},
  2009.

\bibitem{cheng2020learned}
Z.~Cheng, H.~Sun, M.~Takeuchi, and J.~Katto, ``Learned image compression with
  discretized gaussian mixture likelihoods and attention modules,'' in
  \emph{Proc. IEEE Conf. on Computer Vision and Pattern Recognition (CVPR)},
  2020, pp. 7939--7948.

\bibitem{gsplat}
``gsplat,'' \url{https://github.com/nerfstudio-project/gsplat}.

\end{thebibliography}
}

\vfill

\end{document}